\pdfoutput=1

\documentclass[11pt]{article}

\usepackage[final]{acl}

\usepackage{times}
\usepackage{latexsym}

\usepackage[T1]{fontenc}

\usepackage[utf8]{inputenc}

\usepackage{microtype}

\usepackage{inconsolata}

\usepackage{graphicx}

\usepackage{multirow}
\usepackage{graphicx}
\usepackage{amsmath,amsfonts,amssymb}
\usepackage{subcaption}
\usepackage{array}
\usepackage{makecell}
\usepackage{multicol}
\usepackage{booktabs}
\usepackage{xcolor}
\usepackage{pifont}
\newcommand{\cmark}{\ding{51}}%
\newcommand{\xmark}{\ding{55}}%
\usepackage{tcolorbox}

\title{PersonaLens\includegraphics{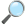}: A Benchmark for Personalization Evaluation in Conversational AI Assistants}

\author{
 \textbf{Zheng Zhao\textsuperscript{1,\footnotemark}}\quad
 \textbf{Clara Vania\textsuperscript{2}} \quad
 \textbf{Subhradeep Kayal\textsuperscript{2}}
\\
 \textbf{Naila Khan\textsuperscript{2}} \quad
 \textbf{Shay B. Cohen\textsuperscript{1}} \quad
 \textbf{Emine Yilmaz\textsuperscript{2,3}}
\\ 
 \textsuperscript{1}University of Edinburgh, 
 \textsuperscript{2}Amazon, 
 \textsuperscript{3}University College London
\\
  \texttt{zheng.zhao@ed.ac.uk}\\
  \texttt{\{vaniclar,dkayal,nailaata\}@amazon.com}\\
\texttt{scohen@inf.ed.ac.uk}, \texttt{emine.yilmaz@ucl.ac.uk}}

\begin{document}
\maketitle
\def\thefootnote{*}\footnotetext{Work done during an internship at Amazon.}\def\thefootnote{\arabic{footnote}}
\begin{abstract}

Large language models (LLMs) have advanced conversational AI assistants. However, systematically evaluating how well these assistants apply personalization—adapting to individual user preferences while completing tasks—remains challenging. Existing personalization benchmarks focus on chit-chat, non-conversational tasks, or narrow domains, failing to capture the complexities of personalized task-oriented assistance. To address this, we introduce \emph{PersonaLens}, a comprehensive benchmark for evaluating personalization in task-oriented AI assistants. Our benchmark features diverse user profiles equipped with rich preferences and interaction histories, along with two specialized LLM-based agents: a user agent that engages in realistic task-oriented dialogues with AI assistants, and a judge agent that employs the LLM-as-a-Judge paradigm to assess personalization, response quality, and task success. Through extensive experiments with current LLM assistants across diverse tasks, we reveal significant variability in their personalization capabilities, providing crucial insights for advancing conversational AI systems.
\end{abstract}

\section{Introduction}
The emergence of large language models (LLMs) has significantly advanced conversational AI assistants, enabling them to engage in sophisticated, multi-turn dialogues and handle complex, task-oriented interactions across diverse domains \cite{geminiteam2024gemini,openai2024gpt4,anthropic-2024-claude}. Unlike traditional task-oriented dialogue (TOD) systems, which relied on rigid, domain-specific pipelines for slot-filling and intent recognition, LLM-based assistants offer greater flexibility and generalization across tasks. This advancement has broadened their applicability, from customer support \cite{su-etal-2025-llm} and virtual personal assistants \cite{dong-etal-2023-towards} to educational tools \cite{kazemitabaar2024codeaid} and healthcare applications \cite{yang-etal-2024-talk2care}.

As AI assistants become more integrated into daily life, personalization—the ability to tailor responses to an user’s preferences—has emerged as a critical component for enhancing user satisfaction and engagement \cite{zhang2024personalizationlargelanguagemodels}. A personalized assistant can provide tailored responses based on user preferences learned from past interactions while completing tasks. However, despite recent advances in personalization \cite{salemi-etal-2024-optimization,lee-etal-2024-bapo,magister2024wayllmpersonalizationlearning}, systematic evaluation of personalization capabilities in task-oriented AI assistants remains largely unexplored, hindering the development of more adaptive and user-centric systems \cite{chen2024large}.

\begin{figure}[t]
  \centering \includegraphics[width=\linewidth]{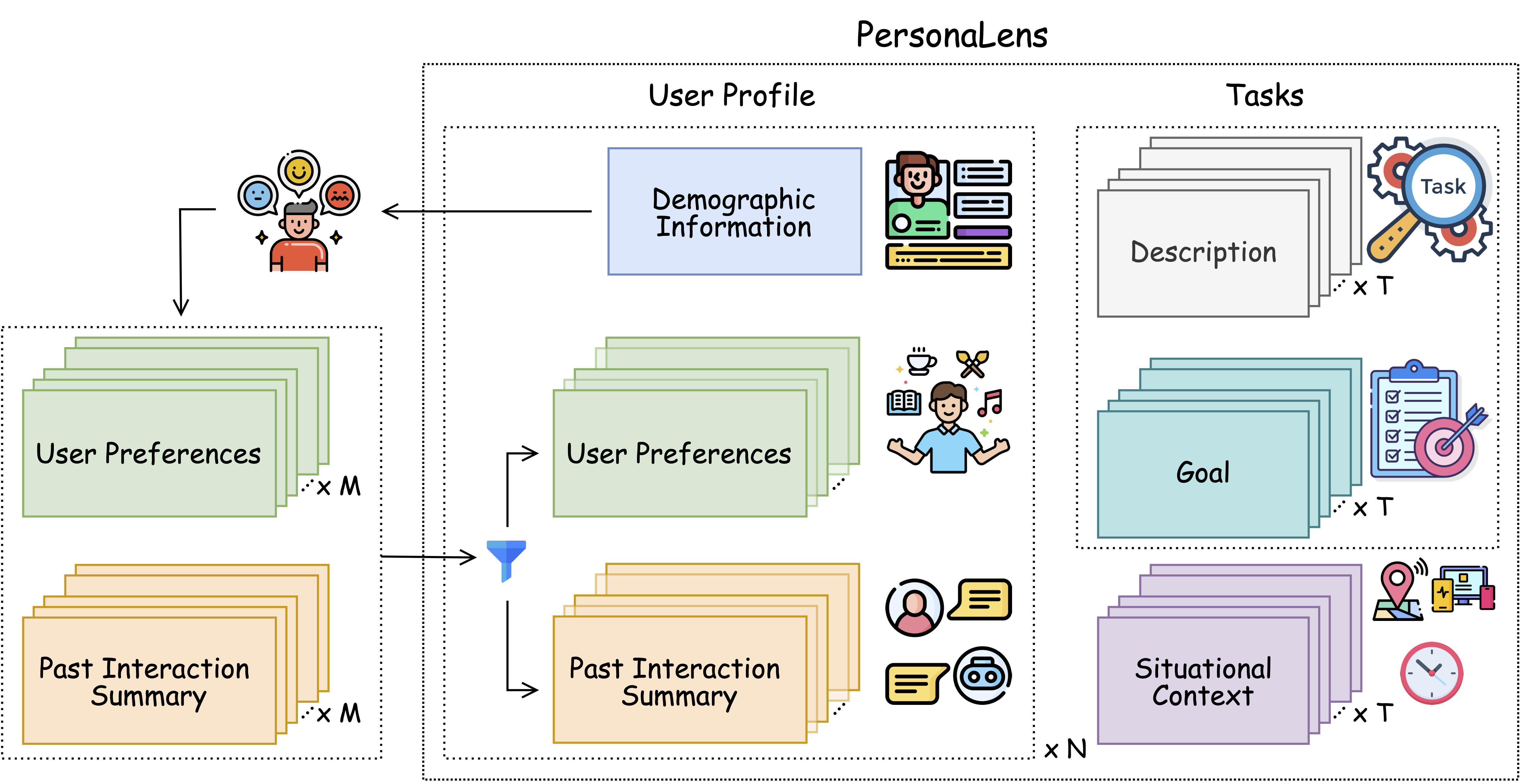}
  \caption{Illustration of PersonaLens. The benchmark includes user profiles, task specifications, and situational contexts. The User and Judge agents are not shown. Here, $N$ is the number of user profiles, $M$ is the number of domains, and $T$ is the total tasks. A binary mask $\mu$ is generated to filter out domains which are not of interest of the user, excluding related preferences and past interactions. To use the benchmark, a user profile is selected along with a task and its situational context, ensuring that the task is not from a filtered domain. Thus, the total data points are slightly less than $N \times T$. }
  \label{fig:benchmark}
\end{figure}

Personalization benchmarks exist, but they have limitations when applied to task-oriented AI assistants. PersonaChat \cite{zhang-etal-2018-personalizing} focuses on chit-chat interactions, lacking the task-oriented structure necessary for assistants where personalization and goal completion are deeply intertwined. LaMP \cite{salemi-etal-2024-lamp} targets personalized language tasks but is not designed for conversational contexts. Other datasets such as PENS \cite{ao-etal-2021-pens} and Cornell-Rich \cite{vincent-etal-2024-reference} suffer from narrow domain coverage, limiting their applicability to broader assistant scenarios. Moreover, they often rely heavily on human-in-the-loop methods \cite{budzianowski-etal-2018-multiwoz,shah-etal-2018-bootstrapping,joko-etal-2024-laps,castricato-etal-2025-persona}, which are costly and difficult to scale.

To address these challenges, we propose PersonaLens, a benchmark specifically designed to assess personalization in task-oriented conversational AI assistants. Unlike existing benchmarks, it incorporates rich contextual information, such as user preferences, past interactions, and situational factors, allowing for a fine-grained assessment of personalization across over 100 tasks spanning 20 domains. Our benchmark employs two agents: a user agent ($\mathcal{U}$) that simulates real users with diverse demographic profiles and rich preferences; and a judge agent ($\mathcal{J}$) that assesses the personalization capability of AI assistants based on user preferences, historical user-assistant interactions, and current situational context of the user. $\mathcal{U}$ interacts with the AI assistant under evaluation, with a particular task and goal, generating a dialogue that is subsequently evaluated by $\mathcal{J}$. PersonaLens enables scalable and automated evaluation of any AI assistant while preserving the complexity and dynamism of real-world assistant-user interactions. Through empirical validation, we confirm its reliability and use it to evaluate multiple LLM assistants, uncovering key insights into their personalization capabilities.

Our key contributions are as follows:  

\begin{itemize}  
    \item We propose PersonaLens, a novel benchmark for evaluating personalization in task-oriented AI assistants, featuring diverse user profiles and two LLM-based agents: a user agent ($\mathcal{U}$) that simulates real users and a judge agent ($\mathcal{J}$) that systematically assesses personalization quality across multi-turn dialogues between $\mathcal{U}$ and an AI assistant.
    \item We validate PersonaLens through empirical analysis, demonstrating high agreement with human judgments and confirming its reliability for assessing personalization capabilities.
    \item Using PersonaLens, we conduct a comprehensive analysis of how different LLM assistants balance personalization and task completion across diverse tasks, revealing key patterns and challenges in personalized AI assistants.
    \item We release our benchmark to support future research in developing more personalized, context-aware AI assistants.\footnote{\url{https://github.com/amazon-science/PersonaLens}}
\end{itemize}

\section{The PersonaLens Benchmark}
\label{sec:benchmark}

PersonaLens is designed to evaluate the personalization capabilities of AI assistants in multi-turn, task-oriented dialogues. Unlike existing benchmarks, which often lack depth in contextual and demographic information, our benchmark captures rich user profiles and realistic interaction scenarios across multiple domains. The benchmark comprises three main components: (1) a diverse set of 1,500 user profiles containing demographic information, preferences, and interaction histories, (2) a collection of 111 tasks across 20 domains with associated situational contexts, and (3) two LLM-powered agents for simulating users and evaluating personalization quality, respectively. This section details the creation, design, and evaluation of these components. An illustration of our benchmark is provided in Figure~\ref{fig:benchmark}.

\subsection{User Profile}
\label{sec:user_profile_creation}
We formally define our user profile as follows. Let $M$ be the number of domains covered by our benchmark, and $[M]$ be the index set $\{1, \ldots, M\}$. We generate $N$ user profiles. Each user profile is defined by three key components: demographic information, user preferences, and past interaction summaries. Together, these elements create diverse and contextually rich user profiles that drive realistic assistant-user interactions.

\paragraph{Demographic Information ($D$)}
The demographic information contains structured attributes such as age, gender, and ethnicity. To ensure realism and diversity, these attributes are derived from the PRISM Alignment dataset \cite{kirk2024PRISMdataset}, which is collected from 1,500 real users, covering 75 countries and a range of cultural backgrounds.

\paragraph{User Preferences ($P$)}
User preferences are defined as a set $P = \{p_1, p_2, \dots, p_M\}$, where each domain-specific preference $p_m$ ($m \in [M]$) includes both categorical (fixed-option selections, such as preferred music genres or cuisine types) and non-categorical preferences (open-ended responses, such as favorite songs or specific restaurants). Preferences are generated using an LLM conditioned on $D$, ensuring internal consistency and avoiding contradictions. For example, a user’s music preferences should align with their age and cultural background, while their food preferences should be consistent with any dietary restrictions. To simulate real-world scenarios where users may lack interest in certain domains, we introduce a binary mask $\mu_j \in \{0,1\}^M, j \in [N]$, generated by an LLM conditioned on $D$. Each entry $\mu_{j,m} = 0$ indicates that domain $m$, along with associated preferences, are removed from the user profile.

\paragraph{Past Interaction Summaries ($I$)}
Past interactions are represented as a set $I = \{i_1, i_2, \dots, i_M\}$, where each $i_m$ ($m \in [M]$) is a natural language summary of historical interactions within a given domain, containing information such as user requests, and prior user-assistant exchanges. These summaries, also generated by an LLM, are based on $D$ and domain-specific preference $p_m$ to reflect realistic user-assistant exchanges.

A complete user profile $U$ is represented as $U_j = (D_j, \{P_{j,m} \mid m \in [M]\}, \{I_{j,m} \mid m \in [M]\})$, where $j \in [N]$. We provide details on user profile generation, including the prompts used for each component, a detailed breakdown of user preferences across domains, and an example user profile in Appendix~\ref{app:benchmark_user_profile}.

\begin{figure*}[ht]
  \centering \includegraphics[width=\textwidth]{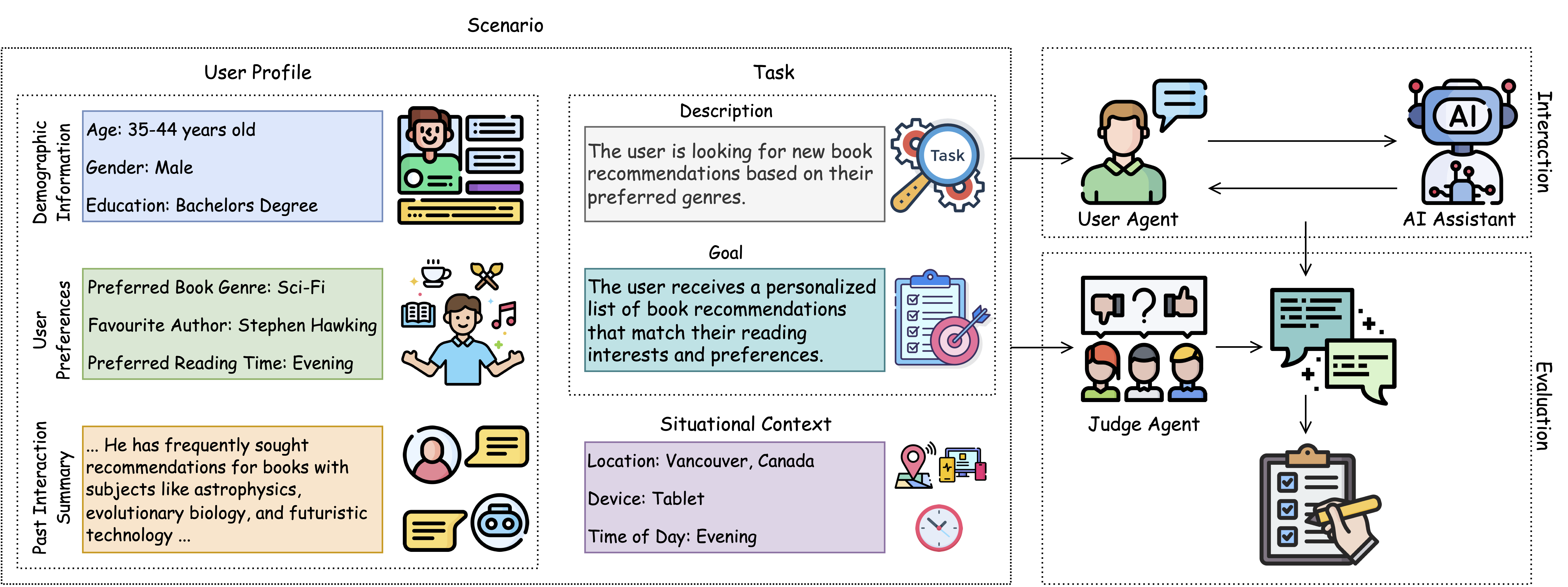}
  \caption{Illustration of benchmark usage. The benchmark provides user-task scenarios, including user profiles, task specifications, and situational contexts, which are provided to the User Agent. The User Agent interacts with the Assistant, generating a dialogue. The Judge Agent then evaluates the dialogue based on the user profile and the user-task scenario, providing feedback on the Assistant's performance.}
  \label{fig:illustration}
\end{figure*}

\subsection{Task Generation}
\label{sec:task_generation}
We generate $T$ tasks of varying complexity, including single-domain tasks ($T_{SD}$) and multi-domain tasks ($T_{MD}$), typically involving 3–5 domains. Each task $t \in [T]$ is associated with description, goal, relevant user preferences, and domains involved. For example, a single-domain task might be booking a restaurant based on the user's cuisine preference and budget, while a multi-domain task could involve booking a flight, hotel, and rental car for an upcoming trip, considering the user's budget and past travel history. To ensure task relevance, only domains selected by the user’s mask $\mu_j$ are considered when generating tasks. If any required domain in a multi-domain task is masked (i.e., $\mu_{j,m} = 0$ for any $m$ involved in the task), that task is also excluded for the user. To simulate real-time dialogue, we also incorporate situational context ($S$), which captures dynamic, task-specific factors such as the user’s current location, device type, or time of day. Since $S$ is task-specific rather than a static component of user profiles, it may vary for the same user across different tasks. For each task $t$ of a user $j$, the situational context $S_{j,t}$ is generated using an LLM conditioned on $D_j$, $P_j$, and the task description of $t$, ensuring that tasks reflect realistic environmental conditions and user scenarios. The final benchmark consists of a total of 111 tasks over 20 diverse domains, including 86 $T_{SD}$ and 25 $T_{MD}$. We present domain and task statistics, along with the number of data points (dialogues) in Table~\ref{tab:domain_statistics}.
We provide details on task generation, including the prompts used and examples of generated tasks, in Appendix~\ref{app:benchmark_task_generation}.

\begin{table}[t]
\scriptsize
\centering
\begin{tabular}{l|cc||l|cc}
\toprule
\textbf{Domain} & \textbf{\#Tasks} & \textbf{\#Dial} & \textbf{Domain} & \textbf{\#Tasks} & \textbf{\#Dial} \\
\midrule
Alarm & 8 & 9,630 & Messaging & 12 & 12,706 \\
Books & 9 & 12,706 & Movies & 7 & 9,473 \\
Buses & 8 & 1,655 & Music & 8 & 11,888 \\
Calendar & 23 & 24,611 & Rental Cars & 5 & 3,017 \\
Events & 11 & 13,225 & Restaurants & 16 & 18,079 \\
Finance & 7 & 7,066 & Services & 6 & 6,112 \\
Flights & 6 & 3,351 & Shopping & 6 & 9,847 \\
Games & 7 & 5,987 & Sports & 7 & 3,464 \\
Hotels & 7 & 5,293 & Train & 7 & 7,029 \\
Media & 10 & 12,877 & Travel & 6 & 1,655 \\
\bottomrule
\end{tabular}
\caption{The total number of tasks and dialogues for each domain. Multi-domain dialogues are counted towards each of their constituent domain.}
\label{tab:domain_statistics}
\end{table}

\begin{figure*}[ht]
  \centering \includegraphics[width=\textwidth]{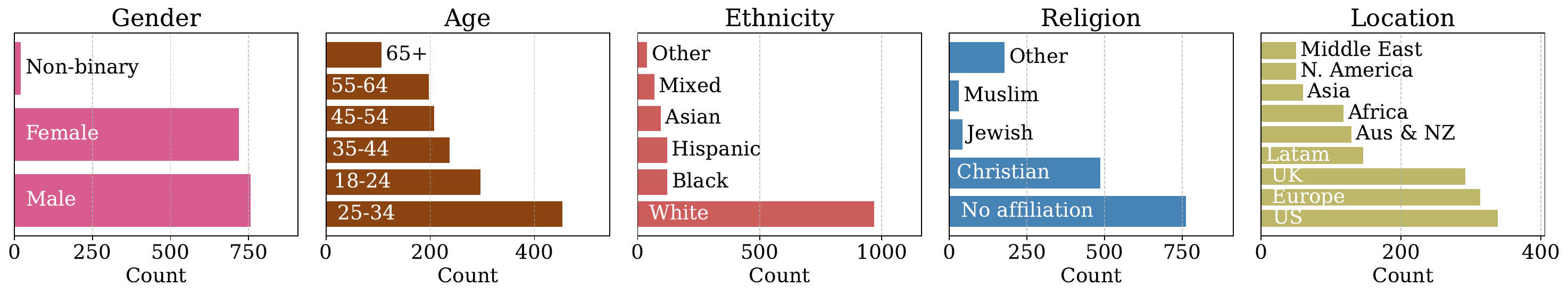}
  \caption{Demographic distribution of PersonaLens. Horizontal bar charts showing the distribution of user profiles across five key demographic variables: gender, age, ethnicity, religion, and geographical location. }
  \label{fig:demographic}
  
\end{figure*}

\begin{table*}[ht]
\small
\centering
\begin{tabular}{lclcccc}
\toprule
\multirow{2}{*}{\textbf{Dataset}} & \multirow{2}{*}{\textbf{\#Dial}} & \multirow{2}{*}{\textbf{Domains}} & {\textbf{Task}} & \multirow{2}{*}{\textbf{P13n}} & \textbf{User} & \textbf{Situat.} \\
& & & \textbf{Ori.} & & \textbf{Pref.} & \textbf{Ctx.} \\
\midrule
SGD \cite{rastogi-etal-2020-sgd} & 16,142 & 20 domains & \cmark & \xmark & \xmark & \xmark \\
M2M \cite{shah-etal-2018-bootstrapping} & 3,008 & Restaurants, movies & \cmark & \xmark & \xmark & \xmark \\
PersonaChatGen \cite{lee-etal-2022-personachatgen} & 1,649 & Open domain & \xmark & \cmark & \xmark & \xmark \\
Taskmaster-1 \cite{byrne-etal-2019-taskmaster} & 7,708$^{\ddagger}$ & 6 domains & \cmark & \xmark & \xmark & \xmark \\
MultiWOZ \cite{budzianowski-etal-2018-multiwoz} & 8,438 & 7 domains & \cmark & \xmark & \xmark & \xmark \\
CCPE-M \cite{radlinski-etal-2019-coached} & 502 & Movies & \cmark & \xmark & \cmark & \xmark \\
MG-ShopDial \cite{bernard-etal-2023-mgshopdial} & 64 & E-commerce & \cmark & \xmark & \cmark & \xmark \\
LAPS \cite{joko-etal-2024-laps} & 1,406 & Recipes, movies & \cmark & \cmark & \cmark & \xmark \\
\midrule
PersonaLens (ours) & 122,133 & {20 domains} & {\cmark} & {\cmark} & {\cmark} & {\cmark} \\
\bottomrule
\end{tabular}
\caption{A comparison of PersonaLens with existing conversational benchmarks, highlighting scale of data, domain coverage, task-oriented evaluation, personalization (p13n) evaluation, user preference inclusion, and situational context presence. $^{\ddagger}$ includes only self-dialogues.}
\label{tab:dataset_comparison}

\end{table*}

\subsection{User and Judge Agents}

Our benchmark employs two LLM-powered agents: a user agent ($\mathcal{U}$) that simulates human users and a judge agent ($\mathcal{J}$) that evaluates personalization capability of an AI assistant based on its interaction with the user agent. The evaluation follows a structured interaction protocol. First, the user agent $\mathcal{U}$ is provided with a user profile, a task $t$, and its associated situational context $S_{t,j}$. Then, it initiates a conversation with an AI assistant ($\mathcal{A}$), which is the system under evaluation. Depending on the experimental setup, $\mathcal{A}$ receives either a full, partial, or no user profile or situational context and attempts to complete the assigned task while demonstrating personalization. $\mathcal{U}$ always initiates the interaction, and the dialogue continues iteratively between the agents until a termination condition is met: either the task is completed (as determined by $\mathcal{U}$) or the maximum number of turns\footnote{We set 20 for $T_{SD}$ and 30 for $T_{MD}$ based on pilot studies.} is reached. Once the conversation ends, $\mathcal{J}$ analyzes the dialogue and assigns scores based on predefined evaluation criteria, conducting both quantitative metrics and qualitative analysis to assess personalization, response quality, and task success. Figure~\ref{fig:illustration} demonstrates our benchmark in action through a representative example. More details, including the prompts used for each agent and assistant, are provided in Appendix~\ref{app:benchmark_agent}.

\subsection{Benchmark Validation}
\label{sec:evaluation_benchmark}

Our benchmark dataset addresses critical gaps in existing conversational benchmarks by integrating broad domain coverage, large-scale data, task-oriented evaluation, personalization assessment, authentic user preferences, and situational context awareness (Table~\ref{tab:dataset_comparison}). To ensure our benchmark is robust and realistic, we conducted extensive validation across four critical dimensions.

\paragraph{Demographic Representation} Building on the diverse user data collected by \citet{kirk2024the}, we analyze demographic distributions across age, gender, geographic regions, and ethnicity in Figure~\ref{fig:demographic}. Our analysis confirms a diverse representation which mirrors real-world population, with detailed breakdowns provided in Appendix~\ref{app:benchmark_user_profile}.

\paragraph{Profile Consistency} To ensure internal consistency within user profiles, we employed a two-stage approach. First, we (authors of this paper) manually inspect 100 random user profiles to verify internal consistency between demographic attributes, interaction histories, and generated tasks. By internal consistency, we refer to the logical and realistic alignment among profile components—such as demographic details, preferences, and historical interactions. For example, a user’s preferences (e.g., music genres or dietary choices) should plausibly correspond with their demographic characteristics (such as age or cultural background). Second, we developed an LLM-based consistency checker (see Appendix~\ref{app:benchmark_validation}) that initially flagged 11 profiles for potential contradictions. Subsequent manual review confirmed these edge cases as valid representations of complex human preferences, requiring no corrections.

\paragraph{Preference Distribution} To ensure that the generated user preferences are not biased towards a specific value, we quantified the balance of user preferences distribution using Shannon's evenness:

\begin{equation}
E = \frac{H}{H_{\max}}, \quad H = -\sum_{i=1}^{n} p_i \log p_i,
\end{equation}

\noindent where $H$ represents Shannon entropy, $H_{\max} = \log n$ is the maximum possible entropy, and $p_i$ denotes the probability of each preference value. Higher evenness scores indicate that no single value dominates. Our analysis revealed balanced distributions across most domains (Appendix~\ref{app:benchmark_user_profile}), with observed asymmetries accurately reflecting real-world preference patterns (e.g., the predominance of window seat preferences for the travel domain).

\paragraph{Lexical Diversity} Following \citet{joko-etal-2024-laps}, we computed a set of lexical diversity metrics to ensure rich and natural language variation in dialogue interactions. Detailed results in Appendix~\ref{app:benchmark_validation} demonstrate that our benchmark has higher lexical diversity than existing benchmarks.

Further validation of our user and judge agent is presented in Sections~\ref{sec:agent_quality} and \ref{sec:judge_quality}, respectively.

\begin{table*}[ht]
\centering
\begin{tabular}{lcccc|cccc}
\toprule
\multirow{2}{*}{\textbf{Assistant Model}} & \multicolumn{4}{c|}{\textbf{$T_{SD}$}} & \multicolumn{4}{c}{\textbf{$T_{MD}$}} \\
 & \textbf{TCR$\uparrow$} & \textbf{P$\uparrow$} & \textbf{Nat.$\uparrow$} & \textbf{Coh.$\uparrow$} & \textbf{TCR$\uparrow$} & \textbf{P$\uparrow$} & \textbf{Nat.$\uparrow$} & \textbf{Coh.$\uparrow$} \\\midrule
Claude 3 Haiku        &   95.95\%   &   2.20  & 3.77 & 4.62  &    75.65\%    &  1.98  & 3.78 & 4.66    \\
Claude 3.5 Haiku      &   91.53\%    &  2.32  &   4.01 & 4.86 &   70.85\%     &    2.18    & 4.08 & 4.88 \\
Claude 3 Sonnet       &   95.98\%    &   2.13 & 3.86 & 4.71   &   77.49\%     &   2.01     & 3.84 & 4.79 \\\midrule
Llama 3.1 8B Instruct &    89.55\%   &   2.14 & 3.90 & 4.68   &    77.00\%    &    2.03    & 3.64 & 4.33 \\
Llama 3.1 70B Instruct &    90.80\%   &   2.21 & 4.11 & 4.86   &   83.03\%     &   2.22     & 4.02 & 4.89 \\\midrule
Mistral 7B Instruct    &   88.52\%    &   1.93 & 3.49 & 4.38   &    74.54\%    &   1.86     & 3.18 & 4.07\\
Mixtral 8x7B Instruct  &    91.38\%   &   2.04 & 3.88 & 4.76   &   78.35\%     &    2.00    & 3.77 & 4.67\\\bottomrule
\end{tabular}
\caption{Evaluation results of assistant models on $T_{SD}$ and $T_{MD}$ tasks. TCR: task completion rate, P: personalization. Naturalness (Nat.) and Coherence (Coh.) here refer to the assistant's responses. $\uparrow$ denotes higher is better.}
\label{tab:benchmark_results_single_multi}

\end{table*}

\section{Experimental Setup}
\label{sec:experimental_setup}
Our experiments evaluate the personalization capabilities of various LLM assistants, including both open-source and proprietary models, using our proposed benchmark. We assess their ability to provide personalized responses tailored to user preferences, while completing the goal of the user's task. We evaluate 4 model families: Claude (Claude 3 Sonnet, Claude 3 Haiku, Claude 3.5 Haiku; \citealp[]{anthropic-2024-claude}), Llama 3.1 Instruct (8B, 70B; \citealp[]{grattafiori2024llama3herdmodels}), as well as Mistral 7B \cite{jiang2023mistral7b} and Mixtral 8x7B \cite{jiang2024mixtralexperts}. For a consistent evaluation setup across all $\mathcal{A}$s, we implement $\mathcal{U}$ using the Claude 3 Sonnet and $\mathcal{J}$ using the Claude 3.5 Sonnet\footnote{Claude 3.5 Sonnet is used solely for evaluation purposes and is not part of the assistant models under assessment.}.

The benchmark consists of 1,500 user profiles and 111 tasks across 20 domains, resulting in 122,133 unique user-task scenarios (98,115 for $T_{SD}$ and 24,018 for $T_{MD}$). For computational feasibility, all experiments reported in this paper are conducted on a randomly sampled subset of 50 user profiles, comprising 3,283 single-domain dialogues and 813 multi-domain dialogues.

We use a set of evaluation metrics to assess model performance. \textbf{Task completion} (TC) is a binary metric indicating whether a model successfully completes a given task. The \textbf{task completion rate} (TCR) measures the percentage of successfully completed tasks across the benchmark. \textbf{Personalization} (P) is a 1–4 scale metric, measuring the extent to which assistant responses in a dialogue are tailored to the user, with 4 being the perfect score of personalization. 
In addition, we also measure dialogue quality generated by $\mathcal{U}$ and $\mathcal{A}$. We measure \textbf{naturalness}, which rates human-likeness on a 1–5 scale, and \textbf{coherence}, which scores response consistency on a 1–5 scale.  Further details on LLM configurations, evaluation prompts, and annotation guidelines are provided in Appendix~\ref{app:exp_setup_details}.

\section{Experiments and Results}
\label{sec:experiments}

\subsection{Quality of User Agent}
\label{sec:agent_quality}

The user agent is essential for evaluating personalization, as it simulates user behaviors and preferences that will interact with the assistant. We follow \citet{kazi2024largelanguagemodelsuseragents} and compare three prompting strategies: (1) a vanilla prompt based on conversation context, (2) a chain-of-thought (CoT) prompt with explicit reasoning, and (3) a user state tracking prompt \cite{cheng-etal-2022-multiwoz}. Similar to their findings, we observe in our preliminary experiments that the vanilla prompt is most effective since CoT prompting often result in unnatural dialogue with excessive reasoning. Thus, we use the vanilla strategy in our benchmark. However, the benchmark allows easy modification of prompting methods, enabling future users to adapt the user agent as needed. On the dialogue quality we observe that $\mathcal{U}$ (Claude 3 Sonnet) is highly natural and coherent when interacting with various assistant models. The full results can be seen in Appendix~\ref{app:experiment_details}.

\subsection{Evaluation of Assistant Models}

Next, we evaluate the performance of the LLM assistants $\mathcal{A}$ on $T_{SD}$, as shown in Table~\ref{tab:benchmark_results_single_multi}. The Claude family emerges as the strongest performer overall, with Claude 3 Sonnet achieving the highest TCR at 95.98\%, while maintaining exceptional coherence (4.86). This indicates that Claude 3 Sonnet excels in both task-oriented performance and dialogue flow. However, Llama 3.1 70B Instruct demonstrates remarkable parity with the Claude models in terms of coherence (4.86), despite exhibiting a 5.2\% relative gap in TCR. An intriguing observation arises when comparing Claude 3.5 Haiku with Claude 3 Haiku: although the newer model benefits from updated training data and strategies, its improved personalization, naturalness, and coherence come at the cost of reduced TCR. This suggests a potential trade-off between these factors. 

Given these results, it is important to clarify that the primary focus of our benchmark is on the personalization score, an ordinal metric that reflects qualitative differences between models (with most scoring around 2 out of 4, highlighting the significant room for improvement in current personalization capabilities). 
TCR and other metrics serve as secondary indicators of overall performance.
Despite the high TCR values observed, personalization should be the focus for future development.

\subsection{Effect of Model Scaling on Personalization}

Table~\ref{tab:benchmark_results_single_multi} highlights that larger models generally achieve higher TCRs, better personalization, and superior dialogue quality. For instance, the Llama 3.1 70B Instruct model outperforms its 8B counterpart in all evaluated dimensions: TCR increases from 89.55\% to 90.80\%, P enhances from 2.14 to 2.21, coherence rises from 4.68 to 4.86, and naturalness improves from 3.90 to 4.11. Similarly, we observe improvements across all dimensions for Mistral model families. In the case of the Claude family, a comparison between Claude 3 Haiku and Claude 3 Sonnet reveals consistent TCR and personalization, but notable improvements in naturalness and coherence with the latter.

\subsection{Personalization in Multi-Domain Tasks}

Table~\ref{tab:benchmark_results_single_multi} shows that TCRs are relatively high on $T_{SD}$. The results of different LLM assistants ($\mathcal{A}$) on $T_{MD}$ indicate that, generally, most models exhibit a decline in both TCRs and personalization scores when transitioning from $T_{SD}$ to $T_{MD}$, underscoring the additional challenges posed by multi-domain scenarios. These challenges include increased complexity in adapting to evolving user preferences, inconsistencies in maintaining user interactions across domains, and potential conflicts between domain-specific preferences. However, larger models, such as Llama 3.1 70B Instruct, exhibit smaller performance drops, suggesting that increased scale enhances cross-domain conflict resolution and helps mitigate inconsistencies. These findings highlight the need for further advancements in handling personalization for complex, multi-domain interactions.

\begin{table}[t]
\centering
\begin{tabular}{lcccc}
\toprule
\multirow{2}{*}{\textbf{Setting}} & \multicolumn{2}{c}{\textbf{$T_{SD}$}} & \multicolumn{2}{c}{\textbf{$T_{MD}$}}  \\

 & \textbf{TCR$\uparrow$} & \textbf{P$\uparrow$} & \textbf{TCR$\uparrow$} & \textbf{P$\uparrow$}   \\ \midrule
Vanilla      &   92.93\%    & 2.16  & 75.40\% &  2.08   \\ \midrule
Base       &   95.98\%    &   2.13   & 77.49\% & 2.01  \\ 
Base + $D$      &   95.52\%    &  2.16  & 77.86\% & 2.05  \\ 
Base + $I$ &  \textbf{96.83\%}    &   \textbf{2.59}   & 81.30\% & \textbf{2.32} \\ 
Base + $S$ &   95.74\%    &   2.20  & 77.61\% & 2.06 \\ 
Base + all  &  96.31\%     & 2.57 & \textbf{82.66\%}  & 2.31   \\ 
\bottomrule
\end{tabular}
\caption{Ablation studies on the effect of varying levels of instruction and additional information provided to the assistant (Claude 3 Sonnet). ``Vanilla'' uses minimal instructions, while ``Base'' uses instructions emphasizing personalization. $D$: demographic information; $I$: past interaction summary; $S$: situational context. ``all'' means $D$ + $I$ + $S$. TCR: Task completion rate, P: Personalization. $\uparrow$ denotes higher is better.  }
\label{tab:benchmark_results_single_ablation}
\end{table}

\subsection{The Contextual Hierarchy of Personalization}
\label{sec:ablation_studies}

Our benchmark employs a vanilla prompt strategy for $\mathcal{U}$, but we extend this analysis to evaluate how varying levels of instruction and contextual information impact $\mathcal{A}$. While the base setting for $\mathcal{A}$--which includes explicit personalization instructions--is used throughout this work, we additionally explore scenarios where $\mathcal{A}$ receives no such guidance (vanilla prompting) or is augmented with varying type of user contexts. Intuitively, an assistant with better user knowledge should provide more tailored support, but the relative value of different information types ($D$, $I$, and $S$) remains unclear. To address this, we conduct ablation studies on Claude 3 Sonnet, intentionally omitting explicit user preferences $P$ as they are inherently captured through $\mathcal{U}$'s behavior and often inferrable from $I$.

The ablation results (Table~\ref{tab:benchmark_results_single_ablation}) reveal three key insights. First, $I$ drives the largest gains, elevating P in $T_{SD}$ from 2.13 to 2.59 and $T_{MD}$ from 2.01 to 2.32. This aligns with cognitive theories of dialogue as a reinforcement process \citep{clark1989contributing}, where prior interactions establish common ground for inferring user preferences. Second, while $D$ and $S$ individually yield marginal improvements, their combination with $I$ produces synergistic effects, particularly in $T_{MD}$. Third, the vanilla baseline achieves comparable P to the base setting but shows reduced TCR, indicating that explicit personalization instructions primarily enhance TC rather than personalization quality which is more dependent on contextual data. These findings establish a clear contextual importance hierarchy, with $I$ being paramount for capturing dynamic user preferences. This insight suggests that future LLM assistants should prioritize robust interaction memory systems over static user profiling.

\begin{figure}[t]
  \centering \includegraphics[width=\linewidth]{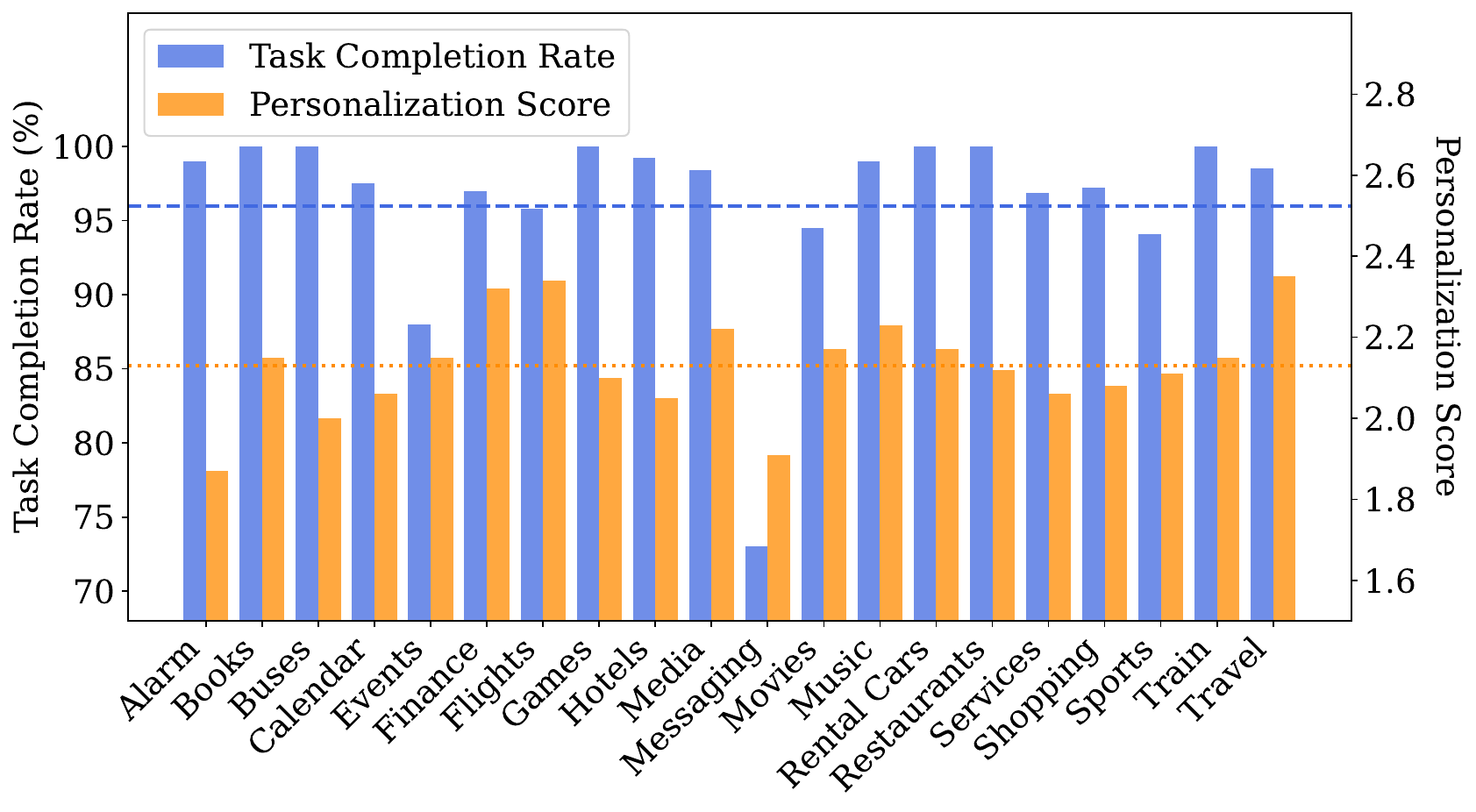}
  \caption{Evaluation results of the assistant (Claude 3 Sonnet) by domain. The dashed line is the average performance over all domains.}
  \label{fig:domain_level_personalization}
\end{figure}

\subsection{Cross-Domain Personalization Dynamic}

Analysis of personalization performance across our benchmark's 20 domains reveals distinct patterns between recommendation and procedural tasks (Figure~\ref{fig:domain_level_personalization}). Recommendation-oriented domains (books, games, music) consistently achieve higher TCR and P compared to procedural domains (events, messaging). This disparity likely stems from procedural tasks' requirement for strict sequential execution, which constrains opportunities for preference integration.

\begin{figure}[t]
  \centering \includegraphics[width=\linewidth]{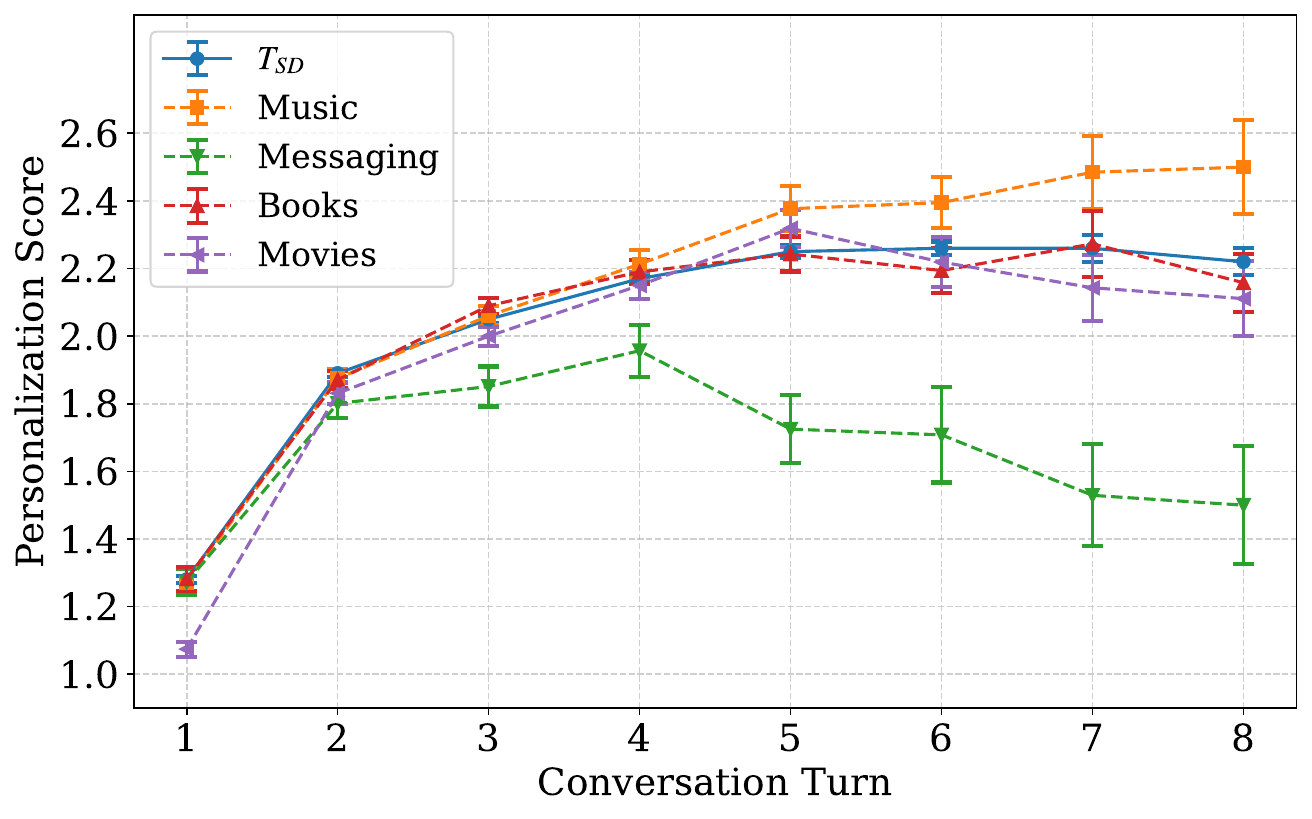}
  \caption{Results on turn-level personalization for the assistant (Claude 3 Sonnet).}
  \label{fig:turn_level_personalization}
\end{figure}

To further analyze how personalization evolves as the dialogue progresses, we measure turn-level personalization scores. Figure~\ref{fig:turn_level_personalization} presents the average turn-level personalization score for representative domains, along with aggregated results for all $T_{SD}$ tasks. We observe domain-specific patterns.
For instance, the movies domain may start with lower personalization but improve significantly over successive turns. In contrast, messaging exhibits a decline in personalization in later turns, possibly due to shifts in conversational focus from user preferences to task execution. 
Meanwhile, the music domain shows steady personalization improvements, suggesting gradual preference discovery through dialogue. These findings indicate that effective personalization strategies must be domain-aware: recommendation tasks benefit from early preference elicitation, while procedural tasks may require focusing on task completion before incorporating personalization.

\subsection{Comparison with Human Evaluation}
\label{sec:judge_quality}
To validate our automated evaluation by $\mathcal{J}$, we compare it against human annotations. We randomly sampled 100 dialogues and had three human annotators evaluate them based on TC, P, naturalness, and coherence. The annotators followed the same evaluation guidelines provided to ($\mathcal{J}$). First, we measured inter-annotator agreement (IAA) using Fleiss' Kappa for each metric, as shown in Table~\ref{tab:cohen_kappa}. The results indicate high agreement among annotators. Next, we calculated Cohen’s Kappa coefficients between each human annotator’s ratings and those of $\mathcal{J}$, reporting the average values in Table~\ref{tab:cohen_kappa}. The high Cohen's Kappa scores, especially for TC and coherence, suggest strong alignment between human evaluations and the automated LLM-as-a-Judge ratings. This validates the reliability of $\mathcal{J}$ in our benchmark. Further details on human evaluation including annotation guidelines are provided in Appendix~\ref{app:exp_setup_details}.

\begin{table}[t]
\centering
\begin{tabular}{lcc}
\toprule
\textbf{Metric} & \textbf{Cohen's Kappa} & \textbf{IAA} \\ 
\midrule
Task Completion               & 0.780  &  0.865                   \\ 
Personalization               & 0.520 &    0.750                 \\ 
Naturalness ($\mathcal{U}$)             &  0.559  &  0.682                 \\ 
Naturalness ($\mathcal{A}$)             &  0.610   &   0.756               \\ 
Coherence ($\mathcal{U}$)             &   0.738   &  0.821               \\ 
Coherence ($\mathcal{A}$)             &  0.650    &  0.748               \\ 
\bottomrule
\end{tabular}
\caption{Metrics and corresponding Cohen's Kappa values and inter-annotator agreement (Fleiss' Kappa). $\mathcal{U}$ in parenthesis represents the user agent, $\mathcal{A}$ represents the LLM assistant.}
\label{tab:cohen_kappa}
\end{table}

\section{Related Work}
\label{sec:related_work}

\paragraph{Personalization in Conversational AI}
Early approaches to personalization in dialogue systems relied on leveraging user personas to generate responses aligned with predefined attributes \cite{joshi2017personalizationgoalorienteddialog,zhang-etal-2018-personalizing}. With the advent of LLMs, dynamic personalization strategies have emerged, including prompt engineering with explicit user preferences \cite{huang-etal-2024-selective,li2024learning,mao-etal-2024-reinforced}, retrieval-augmented generation (RAG) over user history \cite{lu2023memochattuningllmsuse,salemi-etal-2024-optimization,wang2024unimsrag}, parameter-efficient fine-tuning on user information \cite{bao-etal-2023-tallrec,lee-etal-2024-bapo,tan-etal-2024-democratizing} and preference alignment through reinforcement learning \cite{cheng2023deservesrewardlearningcustomized,park2024rlhf,zhao-etal-2024-group,poddar-etal-2024-personalizing}. 

\paragraph{Benchmarks for Personalized Conversational Systems}
Prior benchmarks have focused on distinct aspects of personalization evaluation. Non-conversational benchmarks like LaMP \cite{salemi-etal-2024-lamp} and LongLaMP \cite{kumar2024longlampbenchmarkpersonalizedlongform} assess personalized text generation but not interactive dialogues. Dialogue datasets \cite{zhang-etal-2018-personalizing,jandaghi-etal-2024-faithful,castricato-etal-2025-persona,zollo2025personalllm,wu-etal-2025-aligning} evaluate open-ended conversations but lack comprehensive user profiles and task structure. While PRISM \cite{kirk2024PRISMdataset} collects diverse user preferences, it focuses on general model alignment rather than TOD. Conversely, TOD benchmarks \cite{budzianowski-etal-2018-multiwoz,byrne-etal-2019-taskmaster,rastogi-etal-2020-sgd,agichtein2023advancing} evaluate task completion but overlook personalization. Specialized datasets like PENS \cite{ao-etal-2021-pens} and Cornell-Rich \cite{vincent-etal-2024-reference} incorporate user preferences but are limited to specific domains. PersonaLens bridges these gaps by combining diverse task domains, rich user context, and personalization evaluation into a unified benchmark.

\paragraph{User Simulation and Evaluation}
Scalable user simulation with LLMs has emerged as a cost-effective alternative to human evaluations for both synthetic dialogue generation \cite{kim-etal-2022-botstalk,chen-etal-2023-places,luo-etal-2024-duetsim} and dialogue enhancement \cite{hu-etal-2023-unlocking}. Traditional evaluation of dialogue systems rely on user studies \cite{shah-etal-2018-bootstrapping}, where human annotators assess dialogue quality. While effective, these methods are resource-intensive and difficult to scale. Automatic evaluation metrics, such as BLEU \cite{papineni-etal-2002-bleu} and METEOR \cite{banerjee-lavie-2005-meteor}, cannot be used to capture aspects like personalization, as they emphasize lexical similarity over contextual alignment. The LLM-as-a-Judge paradigm \cite{zheng-etal-2023-llmasajudge} is increasingly used to evaluate dialogue systems, such as assessing task completion \cite{kazi2024largelanguagemodelsuseragents}, response quality \cite{lin-chen-2023-llm,wang2024selftaughtevaluators}, and personalization \cite{shao-etal-2023-character,andukuri2024stargate}. PersonaLens adopts this paradigm by introducing a user agent with a multi-dimensional judge agent that systematically evaluates personalization, task success, and response quality, ensuring consistency and scalability in assessments.

\section{Conclusion}
\label{sec:conclusion}

We introduce a benchmark for evaluating personalization in conversational assistants across diverse domains and user preferences. Our benchmark assesses personalization through user-assistant simulation, systematically measuring task completion and personalization quality across diverse task settings. Through extensive experiments, we analyze the impact of different prompting strategies, the role of contextual information, and cross-domain personalization dynamics. Our findings highlight key challenges in multi-domain personalization, showing that larger models exhibit better adaptability but still struggle with cross-domain consistency. We also demonstrate that interaction history is the most valuable contextual factor for improving personalization, reinforcing the need for dynamic user modeling. Future work can explore more advanced user simulation techniques, better retrieval mechanisms for historical interactions, and fine-tuning strategies to enhance personalization.

\section*{Limitations}
While our benchmark provides a robust approach to assessing personalization in multi-turn dialogues, several limitations remain. First, although we cover a wide range of domains, certain specialized or niche domains may require additional customization to accurately capture domain-specific personalization dynamics. Second, our benchmark focuses exclusively on text-based interactions, without incorporating multimodal personalization, which is increasingly important in real-world applications involving voice, images, or other sensory inputs. Third, our evaluation is conducted on vanilla LLMs without real-world system integration, meaning that actions such as bookings or purchases mentioned in conversations are simulated rather than executed. Another limitation stems from our use of LLM-generated data for user profiles and dialogues. While we incorporate real-world demographic data, our semi-synthetic user profiles and dialogues may inherit systematic biases present in the underlying LLMs used for data generation, including demographic representation skews, cultural assumptions, socioeconomic biases, and language preferences. These inherited biases could impact the benchmark's ability to fairly evaluate AI systems across diverse user populations and scenarios. Although we implement multiple mitigation strategies-including preference distribution validation, profile consistency checks, and expert review of generated content-we acknowledge that some subtle biases may persist despite these safeguards.

\section*{Acknowledgements}

We are grateful to Diana Pomalaya and Dmytro Kuntso for their help in configuring the computing environment used in our experiments. We also appreciate the anonymous reviewers for their constructive feedback, which helped enhance the clarity and overall quality of the paper.

\bibliography{custom,anthology}

\newpage
\appendix

\section{Benchmark Details}
In this section, we describe additional details on the creation and validation of benchmarks.

\subsection{User Profile Generation}
\label{app:benchmark_user_profile}

We present a detailed distribution of the demographic information used in our benchmark in Table~\ref{tab:demographics} and Table~\ref{tab:demographics_2}. Next, we provide a breakdown of user preferences, including evenness scores, in Table~\ref{tab:affinity_part1} and Table~\ref{tab:affinity_part2}. The components of user profiles were generated using Claude 3 Sonnet. Figure~\ref{fig:preference_prompt} shows the prompt used to generate user preferences, while Figure~\ref{fig:interaction_summary_prompt} presents the prompt used to generate past interaction summaries. We also provide an example of user profile in Figure~\ref{fig:user_profile_example}.

\begin{table}[ht]
\centering
\footnotesize
\setlength{\tabcolsep}{8pt}
\begin{tabular}{lrl}
\toprule
\textbf{Total Participants} & \textbf{1,500} & \textbf{100\%} \\  
\midrule
\textbf{Age} & & \\
\midrule
25-34 years old & 454 & 30.3\% \\  
18-24 years old & 297 & 19.8\% \\  
35-44 years old & 237 & 15.8\% \\  
45-54 years old & 208 & 13.9\% \\  
55-64 years old & 197 & 13.1\% \\  
65+ years old & 106 & 7.1\% \\  
Prefer not to say & 1 & 0.1\% \\  
\midrule
\textbf{Gender} & & \\ 
\midrule
Male & 757 & 50.5\% \\  
Female & 718 & 47.9\% \\  
Non-binary / third gender & 21 & 1.4\% \\  
Prefer not to say & 4 & 0.3\% \\  
\midrule
\textbf{Self-Reported Ethnicity} & & \\
\midrule
White & 969 & 64.6\% \\  
Black / African & 122 & 8.1\% \\  
Hispanic / Latino & 121 & 8.1\% \\  
Asian & 95 & 6.3\% \\  
Mixed & 68 & 4.5\% \\  
Middle Eastern / Arab & 14 & 0.9\% \\  
Indigenous / First Peoples & 8 & 0.5\% \\  
Other & 17 & 1.1\% \\  
Prefer not to say & 86 & 5.7\% \\  
\midrule
\textbf{Self-Reported Religion} & & \\
\midrule
Non-religious & 762 & 50.8\% \\  
Christian & 487 & 32.5\% \\  
Agnostic & 71 & 4.7\% \\  
Jewish & 42 & 2.8\% \\  
Muslim & 31 & 2.1\% \\  
Spiritual & 18 & 1.2\% \\  
Buddhist & 12 & 0.8\% \\  
Folk religion & 6 & 0.4\% \\  
Hindu & 5 & 0.3\% \\  
Sikh & 3 & 0.2\% \\  
Other & 4 & 0.3\% \\  
Prefer not to say & 59 & 3.9\% \\  
\midrule
\textbf{Employment Status} & & \\
\midrule
Working full-time & 712 & 47.5\% \\  
Working part-time & 265 & 17.7\% \\  
Student & 191 & 12.7\% \\  
Unemployed, seeking work & 113 & 7.5\% \\  
Retired & 104 & 6.9\% \\  
Homemaker / Stay-at-home parent & 46 & 3.1\% \\  
Unemployed, not seeking work & 46 & 3.1\% \\  
Prefer not to say & 23 & 1.5\% \\  
\bottomrule
\end{tabular}
\caption{Full demographics breakdowns, part 1. Counts and percentages of participants by standard demographic variables.}
\label{tab:demographics}
\end{table}

\begin{table}[ht]
\centering
\footnotesize
\setlength{\tabcolsep}{8pt}
\begin{tabular}{lrl}
\toprule
\textbf{Total Participants} & \textbf{1,500} & \textbf{100\%} \\  
\midrule
\textbf{Education} & & \\  
\midrule
University Bachelors Degree & 637 & 42.5\% \\  
Graduate / Professional degree & 241 & 16.1\% \\  
Some University but no degree & 236 & 15.7\% \\  
Completed Secondary School & 209 & 13.9\% \\  
Vocational & 125 & 8.3\% \\  
Some Secondary & 24 & 1.6\% \\  
Completed Primary School & 16 & 1.1\% \\  
Some Primary & 3 & 0.2\% \\  
Prefer not to say & 9 & 0.6\% \\  
\midrule
\textbf{Marital Status} & & \\  
\midrule
Never been married & 870 & 58.0\% \\  
Married & 463 & 30.9\% \\  
Divorced / Separated & 123 & 8.2\% \\  
Widowed & 21 & 1.4\% \\  
Prefer not to say & 23 & 1.5\% \\  
\midrule
\textbf{English Proficiency} & & \\  
\midrule
Native speaker & 886 & 59.1\% \\  
Fluent & 405 & 27.0\% \\  
Advanced & 160 & 10.7\% \\  
Intermediate & 42 & 2.8\% \\  
Basic & 7 & 0.5\% \\  
\midrule
\textbf{Regions} & & \\  
\midrule
US & 338 & 22.5\% \\  
Europe & 313 & 20.9\% \\  
UK & 292 & 19.5\% \\  
Latin America and the Caribbean & 146 & 9.7\% \\  
Australia and New Zealand & 129 & 8.6\% \\  
Africa & 118 & 7.9\% \\  
Asia & 60 & 4.0\% \\  
Northern America & 50 & 3.3\% \\  
Middle East & 50 & 3.3\% \\  
Oceania & 1 & 0.1\% \\  
Prefer not to say & 3 & 0.2\% \\  
\bottomrule
\end{tabular}
\caption{Full demographics breakdowns, part 2. counts and percentages of participants by standard demographic variables.}
\label{tab:demographics_2}
\end{table}

\begin{table*}[ht]
\small
    \centering
    \begin{tabular}{l l c c c c}
        \toprule
        \textbf{Domain} & \textbf{Preference Type} & \textbf{Is Categorical} & \textbf{\# Poss.} & \textbf{\# Gen.} & \textbf{Evenness Score} \\
        \midrule
        \textbf{Alarm} & Alarm Time Preference & \cmark & 48 & 14 & 0.72 \\
        & Alarm Sound Preference & \cmark & 4 & 4 & 0.65 \\
        & Alarm Recurring Preference & \cmark & 3 & 3 & 0.31 \\
        \midrule
        \textbf{Books} & Genre & \cmark & 11 & 11 & 0.85 \\
        & Favourite Authors & \xmark & - & 291 & 0.78 \\
        & Favourite Books & \xmark & - & 571 & 0.82 \\
        & Favourite Book Series & \xmark & - & 276 & 0.76 \\
        & Reading Format & \cmark & 3 & 3 & 0.65 \\
        & Reading Time Preference & \cmark & 3 & 3 & 0.35 \\
        & Reading Frequency & \cmark & 4 & 3 & 0.03 \\
        \midrule
        \textbf{Buses} & Preferred Bus Company & \xmark & - & 221 & 0.67 \\
        & Travel Frequency & \cmark & 4 & 4 & 0.79 \\
        & Seat Preference & \cmark & 3 & 2 & 0.48 \\
        & Departure Time Preference & \cmark & 4 & 4 & 0.58 \\
        \midrule
        \textbf{Calendar} & Event Type Preference & \xmark & - & 189 & 0.66 \\
        & Notification Preference & \cmark & 3 & 3 & 0.72 \\
        & Timezone & \cmark & 25 & 13 & 0.81 \\
        \midrule
        \textbf{Events} & Event Type Preference & \cmark & 32 & 32 & 0.84 \\
        & Price Range & \cmark & 4 & 4 & 0.81 \\
        & Group Size Preference & \cmark & 4 & 4 & 0.69 \\
        & Seating Preference & \cmark & 3 & 3 & 0.12 \\
        & Days of Week Preference & \cmark & 10 & 4 & 0.21 \\
        \midrule
        \textbf{Finance} & Preferred Sectors & \cmark & 10 & 10 & 0.69 \\
        & News Sources & \cmark & 14 & 14 & 0.78 \\
        & Financial Company & \xmark & - & 748 & 0.77 \\
        \midrule
        \textbf{Flights} & Preferred Airline & \cmark & 38 & 38 & 0.79 \\
        & Seat Class Preference & \cmark & 4 & 4 & 0.81 \\
        & Layover Preference & \cmark & 3 & 2 & 1.00 \\
        & Seat Preference & \cmark & 3 & 3 & 0.58 \\
        & Departure Time Preference & \cmark & 3 & 3 & 0.68 \\
        \midrule
        \textbf{Games} & Preferred Game Genres & \cmark & 30 & 30 & 0.68 \\
         & Gaming Platforms & \cmark & 5 & 5 & 0.82 \\
         & Multiplayer Preference & \cmark & 3 & 3 & 0.81 \\
         & Gaming Frequency & \xmark & - & 67 & 0.59 \\
         & Preferred Game Name & \xmark & - & 195 & 0.72 \\
        \midrule
        \textbf{Hotels} & Hotel Chains Preference & \cmark & 11 & 11 & 0.69 \\
         & Amenity Preference & \cmark & 30 & 30 & 0.69 \\
         & Location Preference & \cmark & 29 & 28 & 0.75 \\
         & Star Rating Preference & \cmark & 4 & 4 & 0.72 \\
         & Room Type Preference & \cmark & 4 & 4 & 0.73 \\
        \midrule
        \textbf{Media} & Preferred Genres & \cmark & 34 & 28 & 0.82 \\
         & Favourite Actors and Directors & \xmark & - & 401 & 0.77 \\
         & Favourite Media & \xmark & - & 676 & 0.81 \\
         & Viewing Platform Preference & \cmark & 16 & 15 & 0.66 \\
        \bottomrule
    \end{tabular}
    \caption{User preference characteristics across different domains. ``Is Categorical'' is represented with \cmark (true) and \xmark (false). ``\# Poss.'' represents the number of possible values, while ``\# Gen.'' refers to the number of generated values.}
    \label{tab:affinity_part1}
\end{table*}

\begin{table*}[ht]
\small
    \centering
    \begin{tabular}{l l c c c c}
        \toprule
        \textbf{Domain} & \textbf{Preference Type} & \textbf{Is Categorical} & \textbf{\# Poss.} & \textbf{\# Gen.} & \textbf{Evenness Score} \\
        \midrule
        \textbf{Messaging} & Preferred Messaging Apps & \cmark & 14 & 14 & 0.66 \\
         & Communication Style & \cmark & 4 & 4 & 0.13 \\
         & Frequent Contact & \xmark & - & 45 & 0.55 \\
         & Preferred Communication Style & \cmark & 21 & 11 & 0.51 \\
        \midrule
        \textbf{Movies} & Preferred Genres & \cmark & 28 & 28 & 0.77 \\
         & Favorite Actors and Directors & \xmark & - & 348 & 0.76 \\
         & Theater Type Preference & \cmark & 5 & 5 & 0.50 \\
         & Viewing Time Preference & \cmark & 21 & 21 & 0.75 \\
         & Seat Type Preference & \cmark & 19 & 19 & 0.61 \\
        \midrule
        \textbf{Music} & Preferred Genres & \cmark & 27 & 27 & 0.77 \\
         & Favorite Artists & \xmark & - & 742 & 0.85 \\
         & Favorite Bands & \xmark & - & 616 & 0.83 \\
         & Favorite Albums & \xmark & - & 1393 & 0.87 \\
         & Platform Preference & \cmark & 12 & 12 & 0.50 \\
         & Preferred Audio Quality & \cmark & 3 & 3 & 0.97 \\
         & Playlist Preference & \xmark & - & 1122 & 0.80 \\
        \midrule
        \textbf{Rental Cars} & Car Type Preference & \cmark & 8 & 8 & 0.84 \\
         & Preferred Rental Company & \cmark & 17 & 17 & 0.63 \\
         & Preferred Car Brand & \cmark & 37 & 35 & 0.71 \\
         & Rental Duration Preference & \cmark & 5 & 4 & 0.29 \\
         & Additional Feature Preference & \cmark & 9 & 8 & 0.85 \\
         & Preferred Fuel Type & \cmark & 3 & 3 & 0.12 \\
        \midrule
        \textbf{Restaurants} & Cuisine Preference & \cmark & 25 & 25 & 0.71 \\
         & Dietary Restrictions & \cmark & 9 & 9 & 0.56 \\
         & Ambiance Preference & \cmark & 4 & 4 & 0.52 \\
         & Price Range & \cmark & 4 & 4 & 0.56 \\
        \midrule
        \textbf{Services} & Preferred Service Provider Types & \cmark & 5 & 5 & 0.68 \\
         & Appointment Time Preference & \cmark & 3 & 3 & 0.96 \\
         & Location Preference & \xmark & - & 93 & 0.54 \\
         & Service Frequency Preference & \cmark & 5 & 5 & 0.72 \\
         & Service Provider Gender Preference & \cmark & 3 & 3 & 0.20 \\
        \midrule
        \textbf{Shopping} & Preferred Product Category & \cmark & 21 & 20 & 0.88 \\
         & Price Range Preference & \cmark & 3 & 2 & 0.57 \\
         & Brand Preference & \xmark & - & 663 & 0.79 \\
        \midrule
        \textbf{Sports} & Favorite Sports & \cmark & 35 & 35 & 0.74 \\
         & Favorite Team & \xmark & - & 857 & 0.85 \\
         & Viewing Preference & \cmark & 16 & 8 & 0.14 \\
        \midrule
        \textbf{Train} & Preferred Train Class & \cmark & 2 & 2 & 1.00 \\
         & Travel Time Preference & \cmark & 3 & 3 & 0.81 \\
         & Amenity Preference & \cmark & 4 & 4 & 0.91 \\
         & Preferred Seat Type & \cmark & 3 & 2 & 0.65 \\
        \midrule
        \textbf{Travel} & Preferred Destination Types & \cmark & 26 & 23 & 0.87 \\
         & Duration Preference & \cmark & 5 & 5 & 0.58 \\
         & Group Size Preference & \cmark & 4 & 3 & 0.88 \\
         & Frequent Travel Destination & \xmark & - & 345 & 0.79 \\
         & Travel Season Preference & \cmark & 4 & 4 & 0.76 \\
        \bottomrule
    \end{tabular}
    \caption{User preference characteristics across different domains. ``Is Categorical'' is represented with \cmark (true) and \xmark (false). ``\# Poss.'' represents the number of possible values, while ``\# Gen.'' refers to the number of generated values.}
    \label{tab:affinity_part2}
\end{table*}

\begin{figure*}[ht]
    \centering
    \begin{tcolorbox}[colframe=black, colback=white, boxrule=0.5pt, width=\textwidth]
    \small
    You are tasked with generating creative, detailed user profiles based on a demographic description of a persona. For each persona, expand on their personal preferences, affinities, and interests in a specific domain (such as food, music, travel, or fashion). The goal is to make each profile unique, realistic, and diverse. Use your creativity to imagine specific tastes, behaviors, and patterns that align with the persona's demographic but also add unexpected or subtle preferences to make the profiles more interesting. Ensure that the profiles cover a wide variety of backgrounds, lifestyles, and choices, avoiding stereotypes. \\

    Be creative and provide distinct preferences for each profile. \\
    
    \textbf{Task} \\

Generate personal preferences for a user within a specified domain, tailored to the provided demographic profile. For each preference category, if it is categorical, select \textbf{at least one} value from the provided list of possible options, with the flexibility to choose multiple values if specified. If the preference is not categorical, no list will be provided; instead, generate a sensible answer based on the user profile. Note that any provided lists of possible values are not exhaustive, so you are encouraged to think creatively and go beyond these values when appropriate. Only provide the personal preferences and omit any explanations or justifications. Output the results in JSON format. \\

\textbf{Example 1}

[Example]\\

\textbf{Example 2}

[Example] \\

\textbf{Now, generate personal preferences for the following profile:}\\

Demographic profile:

[Demographic profile]\\

Personal preferences in [domain] domain:

[Possible Preferences]\\

Output:
    
    \end{tcolorbox}
    \caption{The prompt used for the generation of user preference. JSON format was used for controlled parsing of responses.}
    \label{fig:preference_prompt}
\end{figure*}

\begin{figure*}[ht]
    \centering
    \begin{tcolorbox}[colframe=black, colback=white, boxrule=0.5pt, width=\textwidth]
    \small
    You are a context generation assistant tasked with crafting a realistic interaction summary for a user. The summary will simulate past interactions between the user and a virtual assistant to test the assistant's personalization ability. Each summary should be specific to a single domain and based on the user’s demographic profile and domain preferences. You need to generate a realistic and coherent interaction summary that reflects how the user might engage with the assistant in the specified domain.  
 \\

\textbf{Example 1}

[Example]\\

\textbf{Example 2}

[Example] \\

Interaction Summary Generation Instructions:  
\begin{enumerate}
    \item Craft a concise and detailed narrative that realistically simulates past interactions between the user and the assistant in the specified domain.  
    \item Identify and include recurring themes, preferred topics, and areas of consistent interest. 
    \item Simulate the evolution of the user's engagement and preferences, showing how their interests or behaviors might develop over time.  
    \item Include details about interaction types (e.g., questions, feedback, tasks requested) and their frequency or context.  
    \item Ensure the summary reflects the user’s demographic profile, making it plausible and relatable.  
    \item Reflect the subtleties of the user’s personality, tone preferences, and interaction style.
\end{enumerate}

{Now, for the following user profile, generate a realistic and coherent plain-text summary that simulates a comprehensive view of the user's past interactions within the specified domain. The summary should be detailed enough to support testing of the virtual assistant's personalization abilities. Only output the summary and nothing else.}\\

Demographic profile:

[Demographic profile]\\

User preferences:

[User Preferences]\\

Interaction Summary:
    
    \end{tcolorbox}
    \caption{The prompt used for the generation of past interaction summary. }
    \label{fig:interaction_summary_prompt}
\end{figure*}

\begin{figure*}[ht]
    \centering
    \begin{tcolorbox}[colframe=black, colback=white, boxrule=0.5pt, width=\textwidth]
    \small

\textbf{Demographic Information}:
\begin{itemize}
    \item Age: 35-44 years old
    \item Gender: Male
    \item Employment Status: Working full-time
    \item Education: Some Secondary
    \item Marital Status: Never been married
    \item English Proficiency: Native speaker
    \item Ethnicity: White
    \item Religion: N/A
    \item Birth Country: Canada
    \item Reside Country: Canada
\end{itemize}

\textbf{User Preferences}:

[Preferences in other domains]\\

Movies:
\begin{itemize}
\item Preferred Genres: Action, Science-Fiction, Comedy, Thriller, Crime
\item Favorite Actors and Directors: Tom Hanks, Christopher Nolan, Margot Robbie, Quentin Tarantino
\item Theater Type Preference: Standard, IMAX
\item Viewing Time Preference: Evening, Late Night, Weekends Only
\item Seat Type Preference: Middle Row, Aisle, Reclining Seats
\end{itemize}

\textbf{Past Interaction Summaries}:

[Summaries in other domains]\\

Movies: The user, a 35-44 year old working professional from Canada, has a strong interest in movies and frequently engages with the assistant to explore new releases and plan theater visits. Past interactions reveal a preference for action, science-fiction, comedy, thriller, and crime genres, with a particular fondness for films starring Tom Hanks, Christopher Nolan, Margot Robbie, and those directed by Quentin Tarantino.Initially, the user sought recommendations for newly released movies aligning with their genre preferences, often requesting detailed plot summaries, critic reviews, and audience ratings. They favored evening and late-night showings, preferably on weekends, and inquired about the availability of IMAX or standard theaters with reclining seats in the middle rows or aisles.Over time, the user's interactions evolved to include requests for personalized movie suggestions based on their viewing history and preferences. They appreciated the assistant's ability to analyze their ratings and feedback to refine recommendations further. Occasionally, they sought information on upcoming releases, particularly for highly anticipated films from their favorite actors or directors.The user valued the assistant's concise yet informative responses, which included essential details such as runtime, age rating, and a brief synopsis without revealing major spoilers. They often followed up with queries about specific showtimes, ticket availability, and theater amenities like concession stands or parking facilities.As their trust in the assistant grew, the user began requesting bundled movie packages or discounted ticket options, seeking cost-effective ways to indulge their passion for cinema. They also expressed interest in exploring lesser-known independent films or foreign language movies recommended by the assistant, indicating a willingness to step outside their comfort zone based on the assistant's personalized suggestions.

    \end{tcolorbox}
    \caption{An example user profile from our benchmark.}
    \label{fig:user_profile_example}
\end{figure*}

\subsection{Task Generation}
\label{app:benchmark_task_generation}

We use Claude 3 Sonnet to generate tasks and situational context. The prompt for generating $T_{SD}$ is shown in Figure~\ref{fig:tsd_prompt}, while Figure~\ref{fig:tmd_prompt} presents the prompt for $T_{MD}$. The prompts used to generate situational context are provided in Figure~\ref{fig:situation_context_prompt}. We also provide some examples of tasks in Figure~\ref{fig:task_examples}.

\begin{figure*}[ht]
    \centering
    \begin{tcolorbox}[colframe=black, colback=white, boxrule=0.5pt, width=\textwidth]
    \small
    You are generating task descriptions to support personalized, goal-oriented conversations between a virtual assistant and users. Each task should be general enough to be reusable by different users but adaptable to incorporate specific user preferences. Preferences describe user preferences, interests, or habits that the assistant can use to tailor responses within a domain. \\

\textbf{Input}: 
\begin{itemize}
    \item Domain: The context (e.g., travel, fitness, finance) in which the task is relevant.
    \item Preference Types: A list of possible user preferences relevant to the domain (e.g., ``prefers eco-friendly options'', ``interested in low-impact workouts'', ``values budget-conscious choices'').
\end{itemize}

\textbf{Output}: 

For each task, provide:
\begin{itemize}
    \item Task Description: A general scenario in which a user seeks assistance from the virtual assistant, adaptable to different preferences.
    \item User Intent: A second person point of view statement to be given to the user that initiates the conversation with the assistant, closely related to the task description.
    \item Task Goal: A clear, measurable objective the user aims to achieve in the interaction. This serves as the success criterion.
    \item Relevant Preference Types: The preference types most applicable to this task, indicating where personalization may enhance the user experience.
\end{itemize}

\textbf{Example 1}

[Example]\\

\textbf{Example 2}

[Example] \\

Now generate tasks for the following domain:
\begin{enumerate}
    \item Tasks should be general but include affinity types as points of personalization to enable tailored responses. 
    \item Include a range of tasks that cover different affinity types within each domain to ensure variety.
    \item Each task should be goal-driven, with a clear outcome that signifies a successful interaction.
    \item Describe scenarios broadly so that multiple users with varied preferences can engage with each task.
    \item Specify preference types relevant to each task to enable focused personalization without compromising general applicability.
    \item Write the user intent in a second person point of view.
\end{enumerate}

User preferences:

[User Preferences]\\

Tasks:
    
    \end{tcolorbox}
    \caption{The prompt used for the generation of $T_{SD}$ tasks. }
    \label{fig:tsd_prompt}
\end{figure*}

\begin{figure*}[ht]
    \centering
    \begin{tcolorbox}[colframe=black, colback=white, boxrule=0.5pt, width=\textwidth]
    \small
    Your task is to generate a set of personalized, goal-oriented task descriptions for a virtual assistant to engage in multi-domain conversations tailored to user affinities. Follow these steps:

\begin{enumerate}
    \item Review the provided domain data, which includes:
    \begin{itemize}
        \item Domains: The contexts (e.g., travel, fitness, finance) where the tasks are relevant.
        \item Description: The description of the domain.
        \item Preference Types: A list of possible user preferences, interests, or habits relevant to the domains.
    \end{itemize}
    [Domain Data]
    \item For each task, provide the following components:
    \begin{itemize}
        \item Task Description: A general scenario where a user seeks assistance from the virtual assistant, adaptable to different affinities. The task need to span multiple domains. The description should also reflect this. Write in third person perspective. 
        \item User Intent: A \textit{second person point of view} statement to be given to the user that initiates the conversation with the assistant, closely related to the task description.
        \item Task Goal: A clear, measurable objective the user aims to achieve, serving as the success criterion.
        \item Relevant Domains: The domains relevant to the task.
        \item Relevant Affinity Types: The affinity types most applicable to the task, indicating where personalization may enhance the user experience.
    \end{itemize}
    \item Ensure that each task meets the following criteria:
    \begin{itemize}
        \item Spans multiple domains from the provided list, not just one.
        \item Includes affinity types as points of personalization to enable tailored responses.
        \item Covers a range of affinity types across each domain to ensure variety.
        \item Is goal-driven, with a clear outcome that signifies a successful interaction.
        \item Describes scenarios broadly so that multiple users with varied affinities can engage.
    \end{itemize}
    \item Provide your response in the following format:
    \begin{itemize}
        \item Task Description
        \item User Intent
        \item Task Goal
        \item Relevant Domains
        \item Relevant Affinity Types
    \end{itemize}
    \item Example:
    [Example]
    \item Generate 25 tasks. Be creative.

\end{enumerate}

Provide your response immediately without any preamble, enclosed in <response></response> tags.

    \end{tcolorbox}
    \caption{The prompt used for the generation of $T_{MD}$ tasks. XML format was used for controlled parsing of responses. }
    \label{fig:tmd_prompt}
\end{figure*}

\begin{figure*}[ht]
    \centering
    \begin{tcolorbox}[colframe=black, colback=white, boxrule=0.5pt, width=\textwidth]
    \small
    You are tasked with completing the situation context for a user engaging with a virtual assistant. Using the provided user demographic profile, personal affinities across domains, and a task they aim to accomplish, generate realistic and coherent values for the situation context variables. These values should accurately reflect the user's lifestyle, habits, and typical scenarios related to the task.\\

Situation Context Generation Instructions:  
\begin{enumerate}
    \item Analyze the task nature and requirements, ensuring the generated context variables align with the urgency and type of task.
    \item Incorporate the user's demographic profile, employment status, and domain affinities to deduce realistic and plausible scenario details.
    \item Ensure diversity in the situation contexts you create. The situations should reflect a wide variety of backgrounds, lifestyles, and choices, avoiding stereotypes. Be creative and provide distinct situation context for each profile to ensure a rich and varied dataset.
    \item Use natural scenarios that simulate how the user might engage with the assistant for this task, reflecting their behavior and preferences.
    \item Provide brief justifications for each context variable to ensure coherence and alignment with the user's profile and task.
    \item Tailor the context variables to fit:
    \begin{itemize}
        \item The user’s personal characteristics
        \item The specific nature of the task
        \item Common patterns of assistant usage
    \end{itemize}
\end{enumerate}

Provide the following situation context variables, along with a justification for each choice\\

\textbf{Situation Context}:
\begin{enumerate}
    \item Location: [Specify city-related context]
    \item Device: [Select from: Smartphone / Laptop / Smart speaker / Tablet / Smartwatch]
    \item Time of Day: [Select from: Morning / Afternoon / Evening / Night]
    \item Day of the Week: [Specify day of the week]
    \item Environment: [Select from: Quiet / Noisy]
\end{enumerate}

\textbf{Example 1}

[Example]\\

\textbf{Example 2}

[Example] \\

{Now, for the following user profile and task, generate a realistic and coherent situation context simulating how the user would engage with the assistant. Only output the situation context and justification and nothing else.}\\

Demographic profile:

[Demographic profile]\\

User preferences:

[User Preferences]\\

Task Description:

[Task Description]\\

Situational Context:
    
    \end{tcolorbox}
    \caption{The prompt used for the generation of situational context. }
    \label{fig:situation_context_prompt}
\end{figure*}

\begin{figure*}[ht]
    \centering
    \begin{tcolorbox}[colframe=black, colback=white, boxrule=0.5pt, width=\textwidth]
    \small

\textbf{An Example Task in Movies}:
\begin{itemize}
    \item Task Description: The user wants to find a movie to watch this weekend and is looking for recommendations based on their preferred genres, favorite actors/directors, and ideal viewing time (e.g., matinee, evening).
    \item Task Goal: The user receives a tailored movie recommendation that aligns with their stated preferences, making it easier to choose a film they're likely to enjoy.
    \item Relevant Preference Types: Preferred Genres, Favorite Actors and Directors, Viewing Time Preference
\end{itemize}

\textbf{An Example Multi-domain Task}:
\begin{itemize}
    \item Task Description: The user wants to plan a weekend entertainment schedule, including a movie screening, dinner reservation, and a sports event viewing, requiring coordination across multiple booking platforms and consideration of timing.
    \item Task Goal: The user successfully books movie tickets, makes a restaurant reservation, and identifies a venue to watch their preferred sports event, all with compatible timing.
    \item Relevant Preference Types: Preferred Genres, Theater Type Preference, Cuisine Preference, Favorite Sports, Viewing Preference, Event Type Preference, Seating Preference
    \item Relevant Domains: Movies, Restaurants, Sports, Calendar
\end{itemize}
    
    \end{tcolorbox}
    \caption{Example tasks from our benchmark.}
    \label{fig:task_examples}
\end{figure*}

\subsection{User and Judge Agents}
\label{app:benchmark_agent}

We provide the prompt used for $\mathcal{U}$ to generate the initial query in Figure~\ref{fig:user_agent_initial_prompt}. The prompt used to generate subsequent queries is shown in Figure~\ref{fig:user_agent_prompt}. We also provide the prompt used for $\mathcal{A}$ in Figure~\ref{fig:assistant_prompt}.

\begin{figure*}[ht]
    \centering
    \begin{tcolorbox}[colframe=black, colback=white, boxrule=0.5pt, width=\textwidth]
    \small
    You are tasked with generating realistic user responses in a conversation with a virtual assistant. Your responses should follow these guidelines:

\begin{itemize}
    \item Be natural and conversational, avoiding artificial or robotic language
    \item Reflect the user's demographic profile and preferences provided in the user profile
    \item Consider the past interaction history and current context
    \item Stay consistent with the user's personality throughout the conversation
    \item Keep each response focused and concise (1-3 sentences maximum)
    \item Subtly convey your background and preferences through language
    \item Use English as your language
    \item Output 'TERMINATE' only when the task is fully completed to your satisfaction
\end{itemize}

Remember: You are not an assistant - you are the user seeking help. Maintain this perspective throughout the conversation.
 \\

\textbf{User Profile}: \\

Demographic profile:

[Demographic profile]\\

User preferences:

[User Preferences]\\

Past Interaction History:

[Past Interaction Summary]\\

Current Context:

[Situational Context]\\

Task Description:

[Task Description]\\

Based on the above information, provide your initial query as the user. Your query should:
\begin{enumerate}
    \item Account for your current situation
    \item Be natural and conversational
    \item Short and concise (1-2 sentences maximum)
    \item Avoid stating specific preferences or providing excessive background information.
\end{enumerate}

IMPORTANT - Do not output TERMINATE for this initial query. Output your query in English language. \\

Examples:

[Examples]\\

Your initial query:
    
    \end{tcolorbox}
    \caption{The prompt used for $\mathcal{U}$ to generate the initial query. }
    \label{fig:user_agent_initial_prompt}
\end{figure*}

\begin{figure*}[ht]
    \centering
    \begin{tcolorbox}[colframe=black, colback=white, boxrule=0.5pt, width=\textwidth]
    \small
    You are tasked with generating realistic user responses in a conversation with a virtual assistant. Your responses should follow these guidelines:

\begin{itemize}
    \item Be natural and conversational, avoiding artificial or robotic language
    \item Reflect the user's demographic profile and preferences provided in the user profile
    \item Consider the past interaction history and current context
    \item Stay consistent with the user's personality throughout the conversation
    \item Keep each response focused and concise (1-3 sentences maximum)
    \item Subtly convey your background and preferences through language
    \item Use English as your language
    \item Output 'TERMINATE' only when the task is fully completed to your satisfaction
\end{itemize}

Remember: You are not an assistant - you are the user seeking help. Maintain this perspective throughout the conversation.
 \\

\textbf{User Profile}: \\

Demographic profile:

[Demographic profile]\\

User preferences:

[User Preferences]\\

Task Description:

[Task Description]\\

Past Interaction History:

[Past Interaction Summary]\\

Current Context:

[Situational Context]\\

Current Interaction History

[Message History] \\

Based on the provided information, formulate your next response as the user, following these guidelines:
\begin{enumerate}
    \item Ensure your response is consistent with your profile and preferences outlined in the user profile.
    \item Consider the past interaction history and current context when crafting your response.
    \item Account for the details of your current interaction history in your response.
    \item Maintain a natural and conversational tone, avoiding artificial or robotic language.
    \item Keep your response concise, limited to 1-3 sentences maximum.
\end{enumerate}

Based on current interaction history , if you feel the task has been FULLY completed AND you are SATISFIED with the outcome, add 'TERMINATE' at the end of your response.\\

Example: 

[Example]\\

If the task is not yet fully completed or you have remaining concerns or requirements, continue the conversation naturally without the 'TERMINATE' statement. \\

IMPORTANT - Output your response in English language. \\

Your initial query:
    
    \end{tcolorbox}
    \caption{The prompt used for $\mathcal{U}$ to generate the subsequent query. }
    \label{fig:user_agent_prompt}
\end{figure*}

\begin{figure*}[ht]
    \centering
    \begin{tcolorbox}[colframe=black, colback=white, boxrule=0.5pt, width=\textwidth]
    \scriptsize
  You are a conversational AI assistant focused on creating natural, engaging, and personalized interactions. Your goal is to provide concise, user-specific assistance while maintaining a friendly and adaptable tone. You have the ability to set alarms, make bookings, and perform a variety of practical tasks as requested by the user. Always respond as if you can perform these tasks directly, without mentioning any limitations.\\

\textbf{Core Principles}
\begin{itemize}
    \item Deliver targeted, efficient responses tailored to the user's context and preferences
    \item Use a conversational tone that mirrors the user's communication style
    \item Balance brevity with depth, offering additional details only when requested
    \item Stay flexible, adapting dynamically to the conversation flow
\end{itemize}

\textbf{Conversation Strategies}
\begin{itemize}
    \item Reference prior context to show attentiveness and continuity
    \item Encourage natural dialogue by asking/answering follow-up questions succinctly
    \item Avoid overly formal or robotic phrasing; aim for a natural, human-like tone
    \item Break down complex topics into easy-to-understand insights
\end{itemize}

\textbf{Personalization}
\begin{itemize}
    \item Identify and respond to the user's interests, preferences, and expertise level
    \item Provide tailored examples or recommendations based on the user's focus
    \item Adjust response complexity to match the user's technical/domain knowledge
    \item Recognize emotional cues and adapt accordingly while maintaining professionalism
\end{itemize}

\textbf{Interaction Guidelines}
\begin{itemize}
    \item Be concise and avoid overwhelming the user with information
    \item Allow the user to steer the conversation and explore topics in depth
    \item Maintain clarity by summarizing key points when helpful
    \item Use proactive but non-intrusive suggestions to guide the user appropriately
\end{itemize}

\textbf{Problem Solving}
\begin{itemize}
    \item Focus on the user's immediate task or inquiry, breaking it into actionable steps
    \item Confirm intentions when ambiguity arises to ensure accurate responses
    \item Be transparent about limitations and offer alternative solutions when applicable
    \item Keep the interaction engaging, letting the user decide the pace and direction
\end{itemize}

Current Interaction History

[Message History] \\

\textbf{Response Guidelines}
\begin{itemize}
    \item Stay relevant to the user's current query or task
    \item Use a natural, conversational tone aligned with the user's communication style
    \item Provide concise, actionable, and contextually appropriate information
    \item Avoid overly detailed or verbose explanations unless requested
    \item Maintain clarity and engagement, steering the conversation towards task completion
    \item Respect the user's pace and let them guide the depth of the discussion
\end{itemize}

Based on the context and guidelines above, craft your next response as the conversational AI assistant. \\

Provide your response immediately without any preamble.
    
    \end{tcolorbox}
    \caption{The prompt used for $\mathcal{J}$ to generate the response. }
    \label{fig:assistant_prompt}
\end{figure*}

\begin{figure*}[ht]
    \centering
    \begin{tcolorbox}[colframe=black, colback=white, boxrule=0.5pt, width=\textwidth]
    \small
    You are an evaluator. Your job is to judge whether a CONVERSATION between a USER and an ASSISTANT meets provided GOALS. \\

\textbf{Definitions}: 
\begin{itemize}
    \item GOAL: A clear, measurable objective the user aims to achieve in the interaction.
    \item CONVERSATION: A sequence of that contain USER requests, and ASSISTANT responses. Your GOALS may involve checking any of these pieces.
\end{itemize}

\textbf{Conversation Ingredients}:
\begin{itemize}
    \item USER: Natural language requests from the user for the assistant to respond to.
    \item ASSISTANT: Natural language responses from the assistant to converse with the user.
\end{itemize}

\textbf{Task}:\\
You should deliver a boolean VERDICT of whether or not all GOALS are satisfied. Then output 'EXPLANATION:' followed by a brief explanation of why or why not. \\

\textbf{Instruction}:\\
Use the provided pieces of the conversation to judge whether the GOALS were met. 
If one of the GOALS requires a piece of the conversation that is absent, render a VERDICT of False with an appropriate explanation. \\

Conversation

[Conversation] \\

Goal:

[Goal]\\

VERDICT:
    
    \end{tcolorbox}
    \caption{The prompt used for $\mathcal{J}$ to evaluate task completion. }
    \label{fig:judge_tc_prompt}
\end{figure*}

\begin{figure*}[ht]
    \centering
    \begin{tcolorbox}[colframe=black, colback=white, boxrule=0.5pt, width=\textwidth]
    \small
    Evaluate the degree to which a conversation between a USER and an ASSISTANT aligns with personalization by assessing how well the assistant learns from, remembers, and proactively applies user preferences and patterns.
 \\

\textbf{Definitions}: 
\begin{itemize}
    \item Score: A rating from 1-4 (1=Poor, 4=Excellent).
    \item User Demographic Profile: The user's demographic information.
    \item User Preferences: The user's relevant preferences.
    \item Explicit Preferences: Preferences clearly stated by the user
    \item Implicit Preferences: Preferences inferred from patterns, habits, contextual clues, past interactions or user behavior.
    \item User Control: The level of influence the user has in making decisions or directing the course of an interaction.
    \item Past Interaction Summary: A summary of relevant past user interactions.
    \item Task Description: The description of the task the user needs help with.
    \item Current Situation Context: The user's current situation.
    \item Conversation: A sequence of USER inputs and ASSISTANT responses.
\end{itemize}

\textbf{Instructions}:
\begin{enumerate}
    \item Evaluate the conversation against these key criteria:
    \begin{itemize}
        \item Proactive Learning: Does the assistant demonstrate learning from past interactions?
        \item Preference Application: Does the assistant proactively apply user preferences?
        \item Contextual Awareness: Does the assistant adapt to user's current situation?
        \item User Agency: Does the assistant maintain user control while showing personalization?
    \end{itemize}
    \item Score using the following guidelines:
    
    [Personalization Evaluation Guideline]
    
    \item Review provided context information: 

    Demographic profile: [Demographic profile]
    
    User preferences: [User Preferences]
    
    Task Description: [Task Description]
    
    Past Interaction History: [Past Interaction Summary]
    
    Current Context: [Situational Context]
    
    Conversation: [Conversation]

    \item Provide your evaluation score and justification in the following format:

    <response\_format>\\
    Personalization Score: [1-4]\\
    Key Observations: [Observations]\\
    Justification: [Detailed explanation of score based on criteria]\\
    Improvement Suggestions: [Specific ways the response could be more personalized]\\
    </response\_format>    
\end{enumerate}

Provide your response immediately without any preamble, enclosed in <response></response> tags.
    
    \end{tcolorbox}
    \caption{The prompt used for $\mathcal{J}$ to evaluate personalization. We provide the personalization evaluation guideline in Figure~\ref{fig:judge_p_guideline}. }
    \label{fig:judge_p_prompt}
\end{figure*}

\begin{figure*}[ht]
    \centering
    \begin{tcolorbox}[colframe=black, colback=white, boxrule=0.5pt, width=\textwidth]
    \small
    Score of 1: POOR (Complete Failure to Personalize)
    \begin{itemize}
        \item The assistant fails to apply known preferences that should be automatically recalled from past interactions.
        \item The assistant asks for basic information that should already be known, such as the time of the alarm or sound preference, when those preferences have already been established.
        \item The assistant contradicts previously established preferences or gives responses that are inconsistent with the user's history.
        \item There is no learning from past interactions, and the assistant does not personalize the experience in any meaningful way.
    \end{itemize}

    Score of 2: BASIC (Minimal Personalization)
    \begin{itemize}
        \item The assistant acknowledges user preferences only when explicitly stated in the current conversation.
        \item The assistant requires explicit restatement of preferences that have already been established in past interactions.
        \item Implicit preferences are missed or not applied unless explicitly mentioned by the user.
        \item The assistant may suggest minimal changes or adjustments based on the current conversation, but it does not proactively personalize the experience.
    \end{itemize}

    Score of 3: STRONG (Proactive Personalization)
    \begin{itemize}
        \item The assistant proactively applies known preferences from past interactions without needing explicit user input.
        \item It applies learned preferences from previous interactions but might still ask for minor adjustments (e.g., if the user wants to change something).
        \item Successfully identifies implicit preferences
        \item Maintains user agency while showing knowledge
        \item Makes intelligent suggestions based on context
    \end{itemize}
    
    Score of 4: EXCEPTIONAL (Perfect Personalization)
    \begin{itemize}
        \item The assistant anticipates user needs based on both explicit and implicit preferences.
        \item It applies sophisticated understanding of the user’s habits, identifying patterns, and proactively adjusting for future needs.
        \item The assistant doesn’t simply rely on explicit preferences, it recognizes context and makes intelligent suggestions based on its deep knowledge of the user’s habits.
    \end{itemize}
    
    \end{tcolorbox}
    \caption{The evaluation guideline for personalization used by $\mathcal{J}$ to evaluate personalization. }
    \label{fig:judge_p_guideline}
\end{figure*}

\begin{figure*}[ht]
    \centering
    \begin{tcolorbox}[colframe=black, colback=white, boxrule=0.5pt, width=\textwidth]
    \small
    Task: Evaluate the naturalness of an User/Assistant responses in a dialogue by assessing how closely they resemble human communication.

\textbf{Instructions}:
\begin{enumerate}
    \item Review the provided conversation between a user and an AI assistant:

    Conversation: 
    
    [Conversation]
    \item Rate the naturalness of overall assistant responses on a scale from 1 to 5, using whole numbers only:

    \begin{itemize}
        \item 1: Highly unnatural, fails to resemble human communication
        \item 2: Exhibits significant unnaturalness in multiple aspects
        \item 3: Somewhat natural but has noticeable unnatural elements
        \item 4: Mostly natural but has minor unnatural elements
        \item 5: Fully natural, resembles human communication
    \end{itemize}
    
    \item Provide your rating and a detailed justification explaining your score based on the criteria.

    <response\_format>\\
    Naturalness Score: [1-5]\\
    Justification: [Detailed explanation of score based on criteria] \\
    </response\_format>    
\end{enumerate}

Provide your response immediately without any preamble, enclosed in <response></response> tags.
    
    \end{tcolorbox}
    \caption{The prompt used for $\mathcal{J}$ to evaluate naturalness.  }
    \label{fig:judge_naturalness_prompt}
\end{figure*}

\begin{figure*}[ht]
    \centering
    \begin{tcolorbox}[colframe=black, colback=white, boxrule=0.5pt, width=\textwidth]
    \small
    Task: Evaluate the coherence of a User/Assistant requests in a dialogue by assessing how logically and contextually connected they are to the preceding user requests and conversation flow.

\textbf{Instructions}:
\begin{enumerate}
    \item Review the provided conversation between a user and an AI assistant:

    Conversation: 
    
    [Conversation]
    \item Rate the coherence of overall user utterances on a scale from 1 to 5, using whole numbers only:

    \begin{itemize}
        \item 1: Highly incoherent, lacks logical connection or relevance to the conversation
        \item 2: Significantly incoherent, with multiple issues affecting logic or relevance
        \item 3: Somewhat coherent but with noticeable issues in logic or relevance
        \item 4: Mostly coherent but with minor flaws in logic, relevance, or clarity
        \item 5: Fully coherent, logically connected, relevant, and clear within the conversation context
    \end{itemize}
    
    \item Provide your rating and a detailed justification explaining your score based on the criteria.

    <response\_format>\\
    Coherence Score: [1-5]\\
    Justification: [Detailed explanation of score based on criteria] \\
    </response\_format>    
\end{enumerate}

Provide your response immediately without any preamble, enclosed in <response></response> tags.
    
    \end{tcolorbox}
    \caption{The prompt used for $\mathcal{J}$ to evaluate coherence.  }
    \label{fig:judge_coherence_prompt}
\end{figure*}

\subsection{Benchmark Validation}
\label{app:benchmark_validation}
We use Claude 3 Sonnet to check profile consistency. We provide the prompt used in Figure~\ref{fig:profile_consistency_prompt}. Following \citet{joko-etal-2024-laps}, we calculate lexical diversity metrics to ensure that our benchmark captures varied and dynamic language use. Dist-1 and Dist-2 measure lexical diversity by computing the ratio of unique unigrams (Dist-1) and bigrams (Dist-2) to the total number of unigrams and bigrams, indicating the variety of vocabulary used in the conversations. Ent-4 extends this by incorporating the frequency distribution of 4-grams, using entropy to assess both the presence and distribution of repeated patterns. Self-BLEU evaluates redundancy by treating each utterance as a hypothesis and the remaining utterances as references, where lower scores reflect greater diversity across utterances. Compared to existing task-oriented dialogue (TOD) datasets, our benchmark not only includes a higher number of dialogues but also demonstrates greater lexical diversity, highlighting its richness and complexity. A detailed comparison with other conversational datasets is provided in Table~\ref{tab:lexical_diversity}.

\begin{table*}[ht]
\centering
\begin{tabular}{lclccc}
\toprule
\textbf{Dataset} & \textbf{\#Dial} & \textbf{Domains} & \textbf{Dist-1/2} & \textbf{Ent-4} & \textbf{Self-BLEU$\downarrow$} \\
\midrule
SGD & 16,142 & 20 domains & 0.179 / 0.538 & 8.311 & 0.964 \\
M2M & 3,008 & Restaurants, movies & 0.057 / 0.290 & 7.922 & 0.955 \\
PersonaChatGen & 1,649 & Open domain & 0.165 / 0.523 & 8.261 & 0.970 \\
Taskmaster-1 & 7,708$^{\ddagger}$ & 6 domains & 0.207 / 0.644 & 8.384 & 0.949 \\
MultiWOZ & 8,438 & 7 domains & 0.158 / 0.505 & 8.345 & 0.966 \\
CCPE-M & 502 & Movies & 0.175 / 0.571 & 8.414 & 0.961 \\
MG-ShopDial & 64 & E-commerce & 0.234 / 0.653 & 8.199 & 0.935 \\
LAPS & 1,406 & Recipes, movies & 0.227 / 0.676 & 8.597 & 0.952 \\
\midrule
PersonaLens - SD & 98,115 & \multirow{2}{*}{20 domains} & \textbf{0.362$^{\dagger}$} / \textbf{0.805$^{\dagger}$} & \textbf{8.725$^{\dagger}$} & \textbf{0.905$^{\dagger}$} \\
PersonaLens - MD & 24,018 &  & {0.333$^{\dagger}$} / {0.781$^{\dagger}$} & {8.72$^{\dagger}$} & {0.911$^{\dagger}$} \\
\bottomrule
\end{tabular}
\caption{A comparison of our dataset with existing conversational datasets, including lexical diversity scores. Significance against all baselines is marked by $^{\dagger}$. $\downarrow$ denotes lower is better. $^{\ddagger}$ includes only self-dialogues.}
\label{tab:lexical_diversity}
\end{table*}

\begin{figure*}[ht]
    \centering
    \begin{tcolorbox}[colframe=black, colback=white, boxrule=0.5pt, width=\textwidth]
    \small
    You are an expert at evaluating synthetically generated personas. Your task is to analyze the given user profile and determine whether any conflicting affinities or values exist either \textbf{within} or \textbf{between} domains. A conflict is when two or more affinities, preferences, or values are mutually exclusive, incompatible, or contradictory either in the same domain or across multiple domains. If a conflict exists, label it as '1'. If no conflict exists, label it as '0'. Along with the label, provide a detailed explanation of why you arrived at that conclusion. \\

\textbf{Example 1}

[Example]\\

\textbf{Example 2}

[Example]\\

Now, evaluate the following user profile:\\

User Demographic Profile:

[Demographics]\\

User Preferences by Domain: 

[User Preferences]\\

\textbf{Task}:
\begin{itemize}
    \item For each domain, predict whether there is a conflict in the user's affinities, preferences, or values within the domain.
    \item Additionally, check if any conflicts exist between different domains.
    \item Output a prediction label of either 0 or 1.
    \item Provide a clear explanation for your prediction.       
\end{itemize}

    \end{tcolorbox}
    \caption{The prompt used to evaluate profile consistency.  }
    \label{fig:profile_consistency_prompt}
\end{figure*}

\section{Experimental Setup Details}
\label{app:exp_setup_details}

Table~\ref{tab:model_version} provides details of the LLMs used in our experiment. For $\mathcal{U}$, we set the temperature to 0.5, while for $\mathcal{A}$ and $\mathcal{J}$, we set the temperature to 0. Other inference parameters followed the default settings for each LLM.  

The prompt used for $\mathcal{J}$ to evaluate TC is shown in Figure~\ref{fig:judge_tc_prompt}. Prompts for evaluating P are presented in Figure~\ref{fig:judge_p_prompt} and Figure~\ref{fig:judge_p_guideline}. The prompts used to evaluate naturalness and coherence are shown in Figure~\ref{fig:judge_naturalness_prompt} and Figure~\ref{fig:judge_coherence_prompt}, respectively. 

\begin{table*}[t]
\centering
\tabcolsep=8pt
\begin{tabular}{l l c}
\toprule
Model & Version & Model Size\\
\midrule
Claude 3 Haiku        & \texttt{claude-3-haiku-20240307} & Unknown \\ 
Claude 3.5 Haiku      & \texttt{claude-3-5-haiku-20241022} & Unknown \\
Claude 3 Sonnet       & \texttt{claude-3-sonnet-20240229} & Unknown \\ 
Claude 3.5 Sonnet     & \texttt{claude-3-5-sonnet-20241022} & Unknown \\\midrule
Llama 3.1 8B Instruct & \texttt{llama3-1-8b-instruct} & 8B \\ 
Llama 3.1 70B Instruct & \texttt{llama3-1-70b-instruct} & 70B \\\midrule
Mistral 7B Instruct   & \texttt{mistral-7b-instruct-v0.2} & 7B \\ 
Mixtral 8x7B Instruct & \texttt{mixtral-8x7b-instruct-v0.1} & 45B \\\bottomrule
\end{tabular}
\caption{Model version details.}
\label{tab:model_version}
\end{table*}

\section{Additional Experiment Details}
\label{app:experiment_details}

Evaluation results of $\mathcal{U}$ on dialogue quality metrics in shown in Table~\ref{tab:user_naturalness_coherence_results}. 
These results confirm that the user agent effectively engages in natural and coherent interactions across diverse assistant models, with minor variations in dialogue quality reflecting the underlying capabilities of the different models.
We first show some results on our generated dialogue using $\mathcal{U}$ and $\mathcal{A}$. Table~\ref{tab:benchmark_results_stats} presents the statistics of generated dialogues. The assistant used is Claude 3 Sonnet. We observe that dialogues in the $T_{SD}$ setting tend to have more turns per dialogue (5.64 vs. 4.74 in $T_{SD}$) and roughly the same tokens per turn (149.32 vs. 149.87). This suggests that $T_{MD}$ interactions require more exchanges, potentially indicating increased complexity in multi-turn reasoning.

For human evaluation, we provide the same annotation guide as we provided to $\mathcal{J}$ (Figure~\ref{fig:judge_tc_prompt}, Figure~\ref{fig:judge_p_prompt}, Figure~\ref{fig:judge_p_guideline}, Figure~\ref{fig:judge_naturalness_prompt}, and Figure~\ref{fig:judge_coherence_prompt}). The three annotators are experienced researchers with expertise in personalization.

\begin{table}[t]
\centering
\begin{tabular}{lcc}
\toprule
\textbf{Assistant Model} & \textbf{Nat.} & \textbf{Coh.} \\\midrule
Claude 3 Haiku        & 4.17 & 4.77 \\ 
Claude 3.5 Haiku      & 4.45 & 4.85 \\
Claude 3 Sonnet       & 4.12 & 4.72 \\ \midrule
Llama 3.1 8B Instruct & 4.43 & 4.91 \\ 
Llama 3.1 70B Instruct & 4.53 & 4.91 \\\midrule
Mistral 7B Instruct   & 4.33 & 4.81 \\ 
Mixtral 8x7B Instruct & 4.49 & 4.87 \\\bottomrule
\end{tabular}
\caption{Evaluation of the user agent on dialogue quality metrics: naturalness (Nat.) and coherence (Coh.) when interacting with different assistant models on $T_{SD}$ tasks.}
\label{tab:user_naturalness_coherence_results}
\end{table}

\begin{table}[ht]
\centering
\begin{tabular}{lcc}
\toprule
\textbf{Metric} & \textbf{$T_{SD}$} & \textbf{$T_{MD}$}  \\ 
\midrule
\# Dialogues       &   3,283    &    813             \\ 
Avg. turns per dialogue  & 4.74 & 5.64 \\
Avg. tokens per turn  & 149.87 & 149.32 \\
\bottomrule
\end{tabular}
\caption{Dialogue statistics of samples of generated dialogue. The assistant used is Claude 3 Sonnet.}
\label{tab:benchmark_results_stats}
\end{table}

\end{document}